\documentclass[10pt,twocolumn,letterpaper]{article}

\usepackage{iccv}
\usepackage{times}
\usepackage{epsfig}
\usepackage{graphicx}
\usepackage{amsmath}
\usepackage{mathrsfs}
\usepackage{amssymb}
\usepackage{url}
\usepackage{caption}
\usepackage{enumitem}
\usepackage{mathrsfs}
\usepackage{algorithm}
\usepackage{algpseudocode}
\usepackage{dsfont}
\usepackage[pagebackref=false,breaklinks=true,letterpaper=true,colorlinks,urlcolor=blue,citecolor=blue,linkcolor=blue,bookmarks=false]{hyperref}
\usepackage{breakurl}
\usepackage{multirow}
\usepackage{booktabs}

\iccvfinalcopy 

\def\iccvPaperID{1118} 

\newcommand{\figdir}{figures}

\usepackage{color}

\newcommand{\ryn}[1]{{\color{black}{#1}}}

\ificcvfinal\fi
\begin{document}

\title{CREST: Convolutional Residual Learning for Visual Tracking}

\author{Yibing Song$^1$, Chao Ma$^2$, Lijun Gong$^3$, Jiawei Zhang$^1$, Rynson \ryn{W.H.} Lau$^1$, and Ming-Hsuan Yang$^4$\\\\
$^1$City University of Hong Kong, \ryn{Hong Kong}~~~~~~~~~~$^2$The University of Adelaide, \ryn{Australia}\\
~~~~~~~~$^3$SenseNets Technology Ltd, \ryn{China}~~~~~~~~~~~~~~~$^4$University of California, Merced, \ryn{U.S.A.}\\
}

\maketitle

\begin{abstract}
Discriminative correlation filters (DCFs) have been shown to perform superiorly in visual tracking. They only need a small set of training samples from the initial frame to generate an appearance model. However, existing DCFs learn the filters separately from feature extraction, and update these filters using a moving average operation with an empirical weight. These DCF trackers hardly benefit from the end-to-end training. In this paper, we propose the CREST algorithm to reformulate DCFs as a one-layer convolutional neural network. Our method integrates feature extraction, response map generation as well as model update into the neural networks for an end-to-end training. To reduce model degradation during online update, we apply residual learning to take appearance changes into account. Extensive experiments on the benchmark datasets demonstrate that our CREST tracker performs favorably against state-of-the-art trackers. \footnote{More results and code are provided on the authors' webpages.}
\end{abstract}

\section{Introduction}
Visual tracking has various applications ranging from video surveillance, human computer interaction to autonomous driving. The main difficulty is how to utilize the extremely limited training data (usually a bounding box in the first frame) to develop an appearance model robust to a variety of challenges including background clutter, scale variation, motion blur and partial occlusions. Discriminative correlation filters (DCFs) have attracted an increasing attention in the tracking community \cite{bolme-cvpr10-mosse,Danelljan-iccvw15-DeepSRDCF,chao-iccv15-HCF}, due to the following two important properties. First, since spatial correlation is often computed in the Fourier domain as an element-wise product, DCFs are suitable for fast tracking. Second, DCFs regress the circularly shifted versions of input features to soft labels, i.e., generated by a Gaussian function ranging from zero to one. In contrast to most \ryn{existing} tracking-by-detection approaches \cite{kalal-pami12-tld,bai-iccv13-randomized,hare-pami16-struck,ning-cvpr16-dlsvm} that generate sparse response scores over sampled locations, DCFs always generate dense response scores over all searching locations. With the use of deep convolutional features \cite{krizhevsky-nips12-imagenet}, DCFs based tracking algorithms \cite{chao-iccv15-HCF,Danelljan-iccvw15-DeepSRDCF,martin-eccv16-beyond} have achieved state-of-the-art performance on recent tracking benchmark datasets \cite{wu-cvpr13-otb,wu-pami15-otb,kristan-eccvw16-vot}.

\renewcommand{\tabcolsep}{.1pt}
\def\swtwo{0.48\linewidth}
\def\swone{0.9\linewidth}
\begin{figure}[t]
\begin{center}
\begin{tabular}{cc}
\vspace{-1mm}\includegraphics[width=\swtwo]{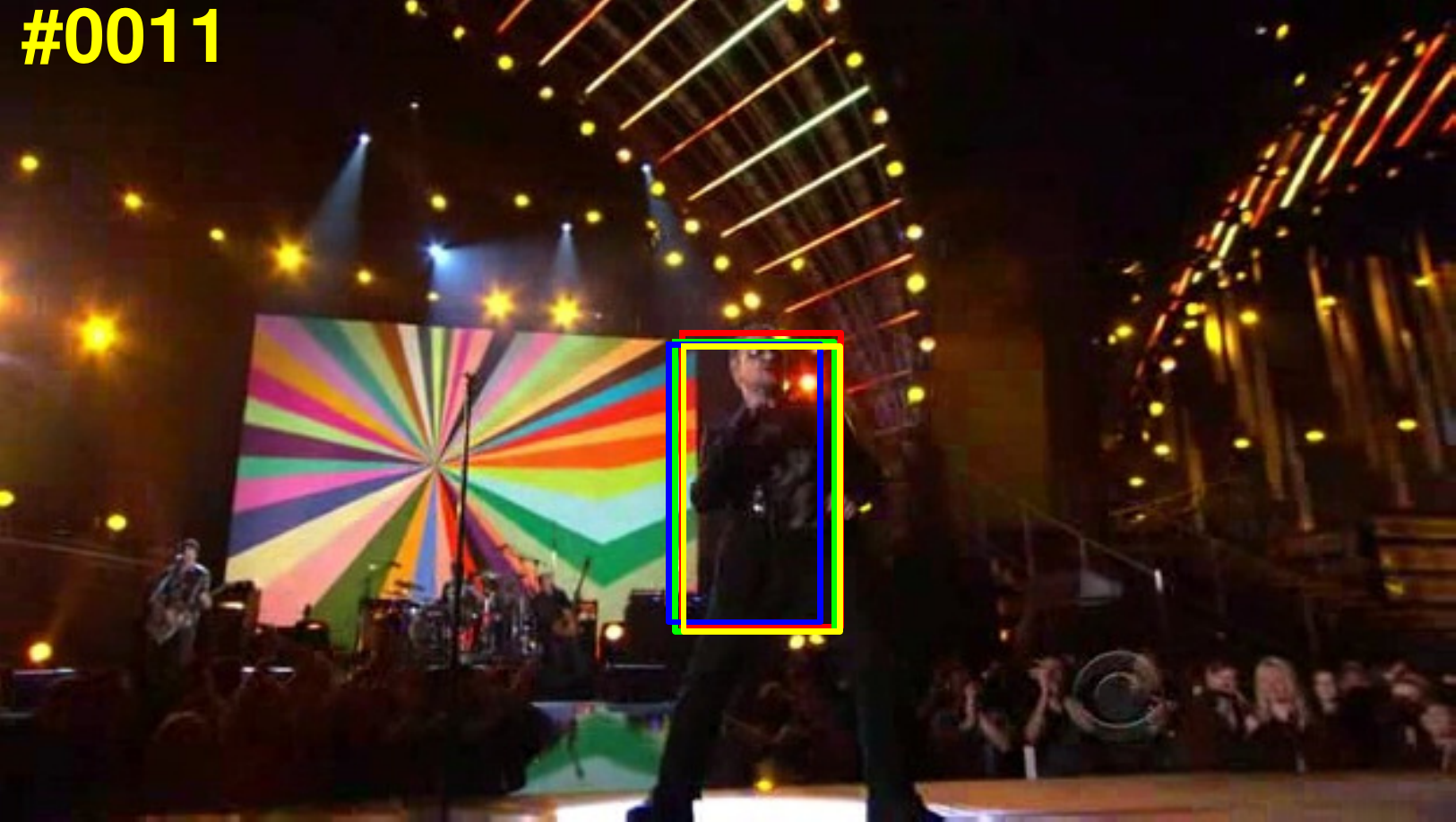}&
\includegraphics[width=\swtwo]{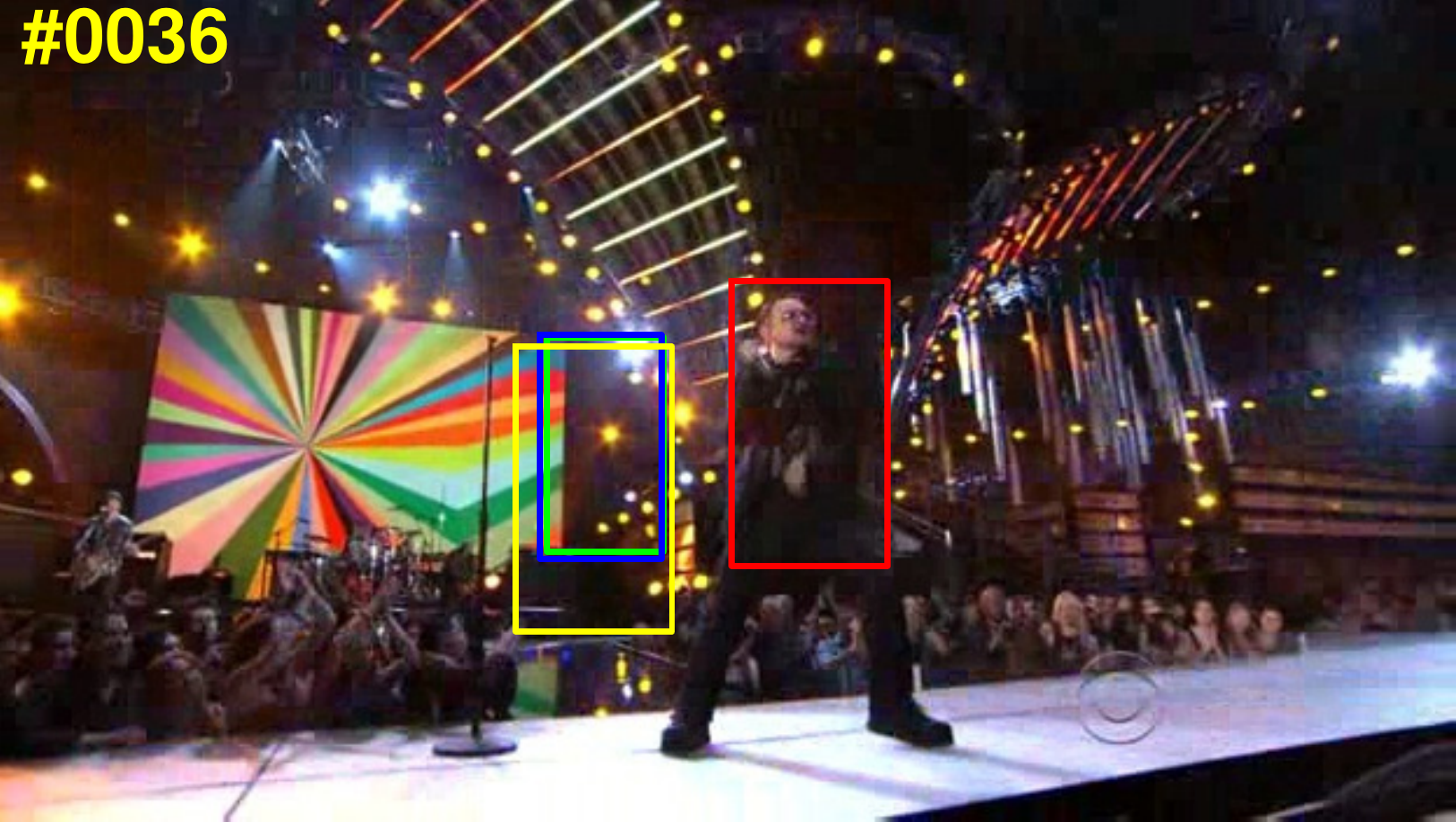}\\
\vspace{-1mm}\includegraphics[width=\swtwo]{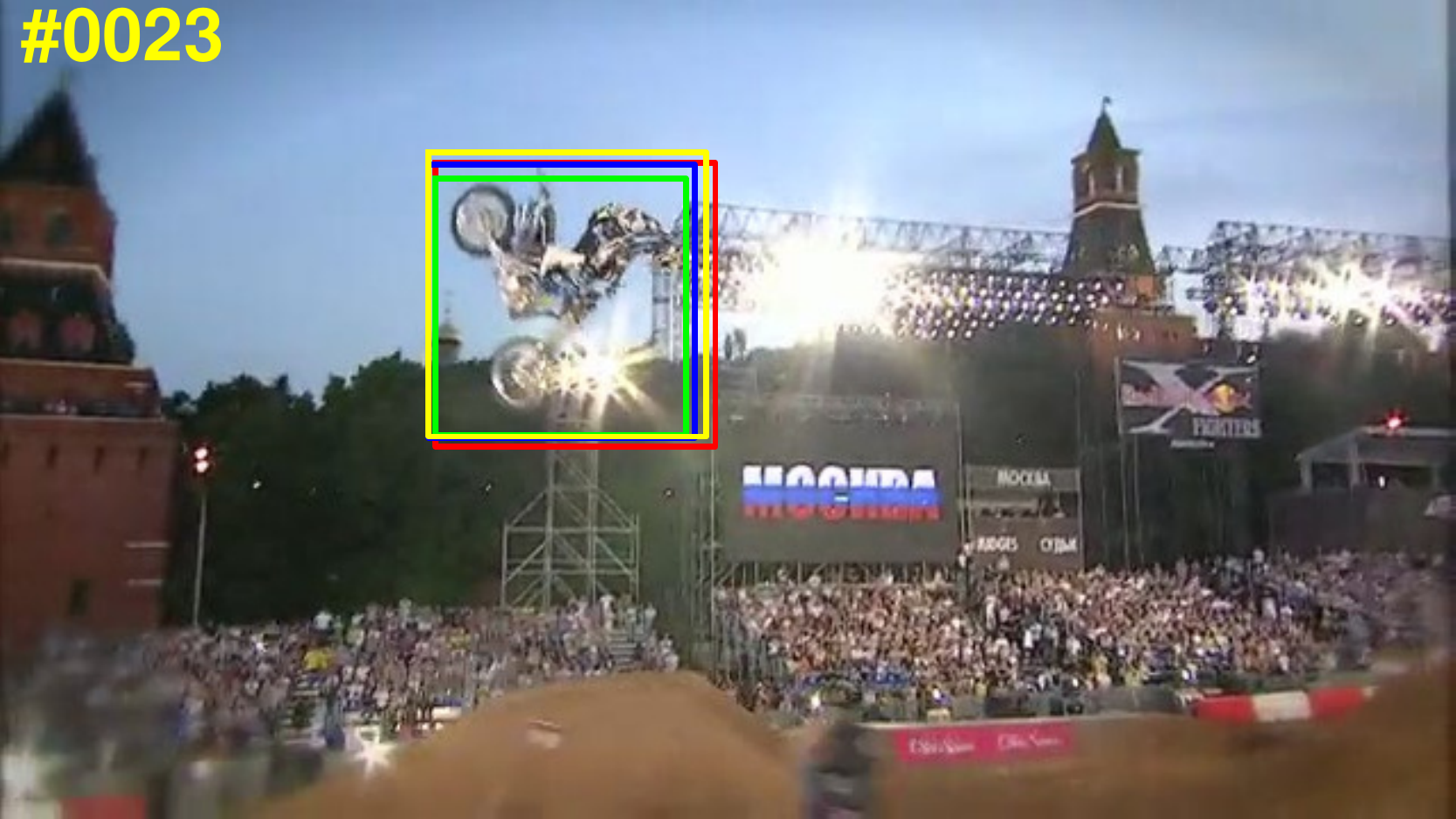}&
\includegraphics[width=\swtwo]{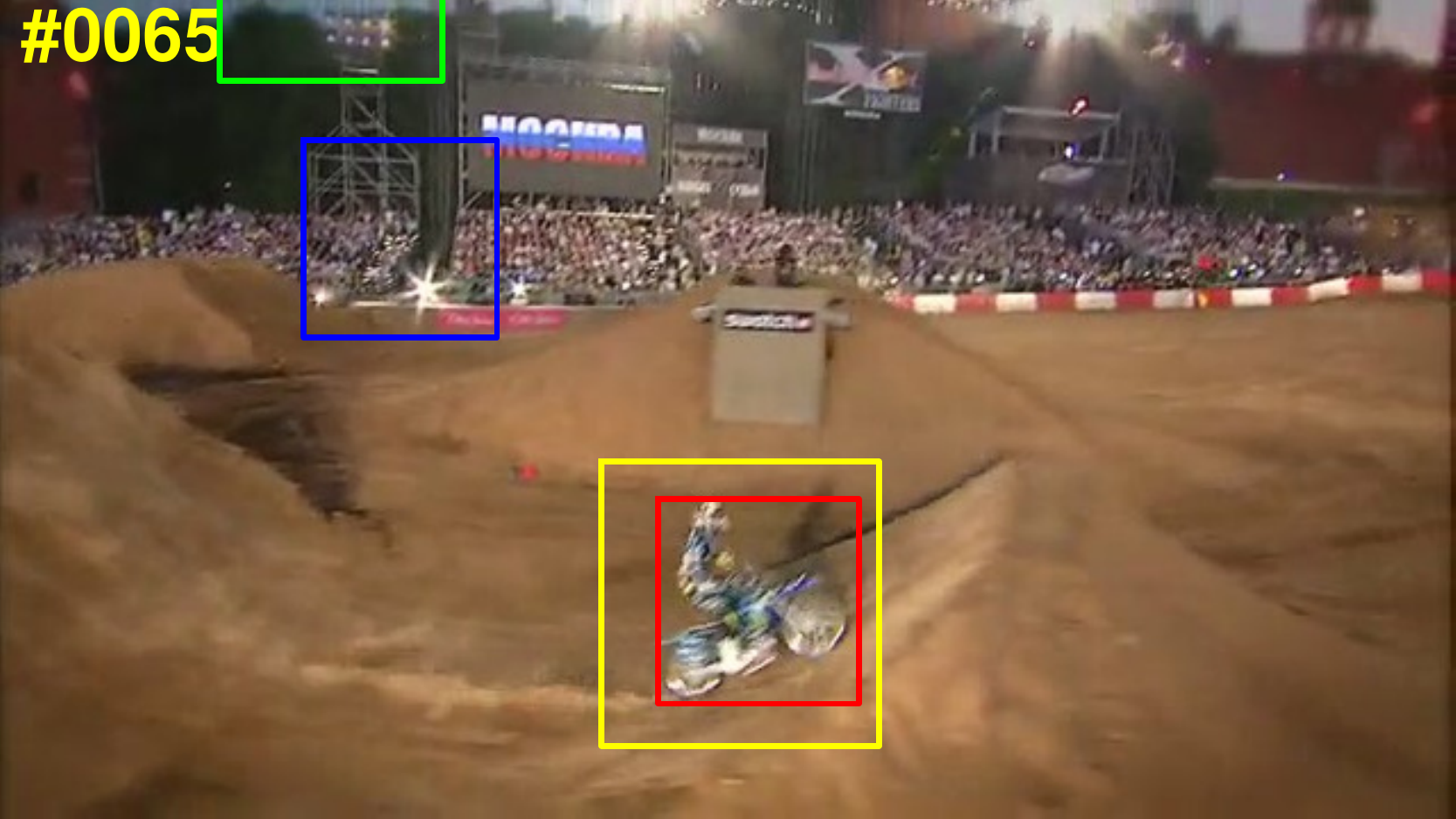}\\
\vspace{-1mm}\includegraphics[width=\swtwo]{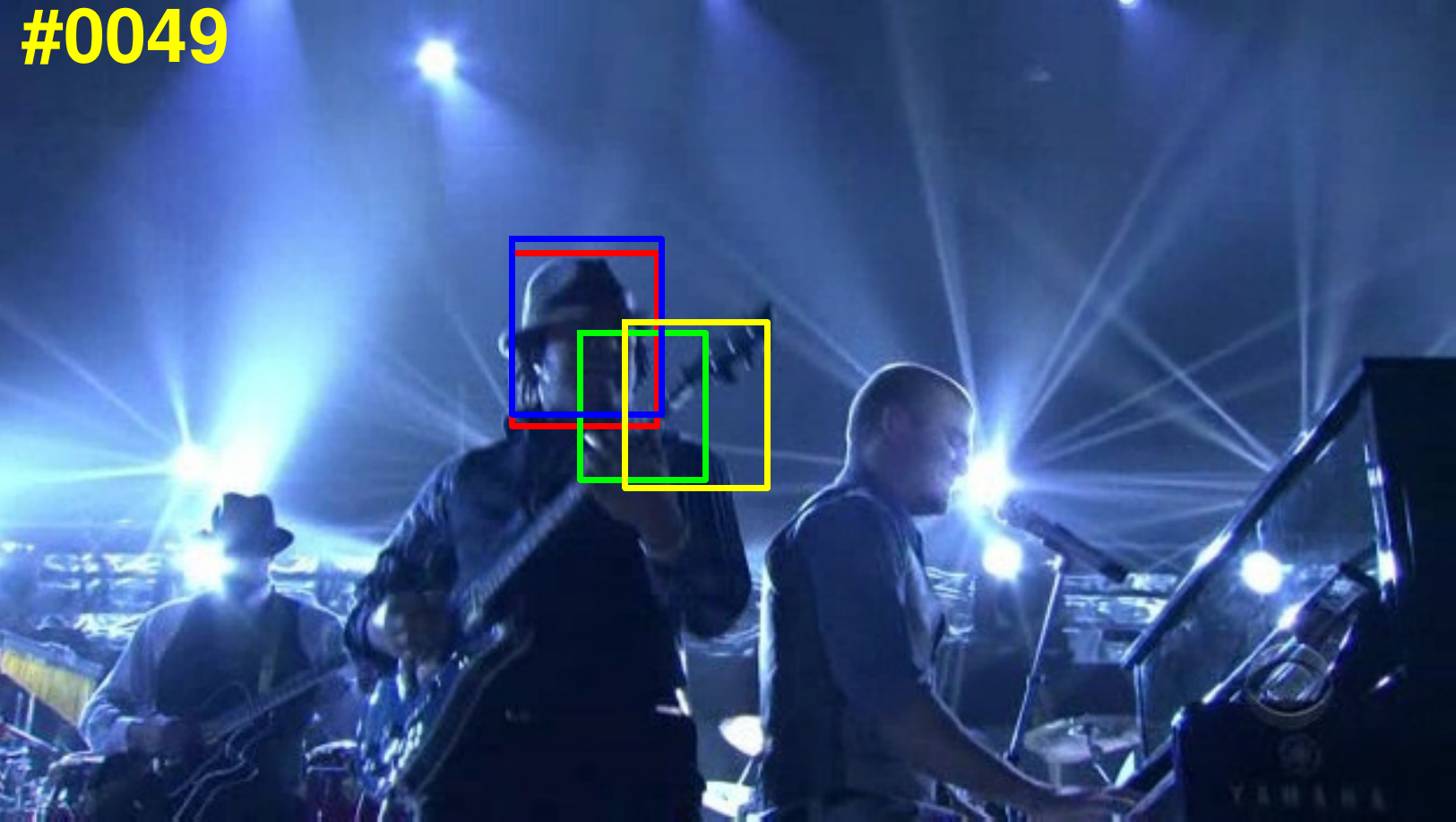}&
\includegraphics[width=\swtwo]{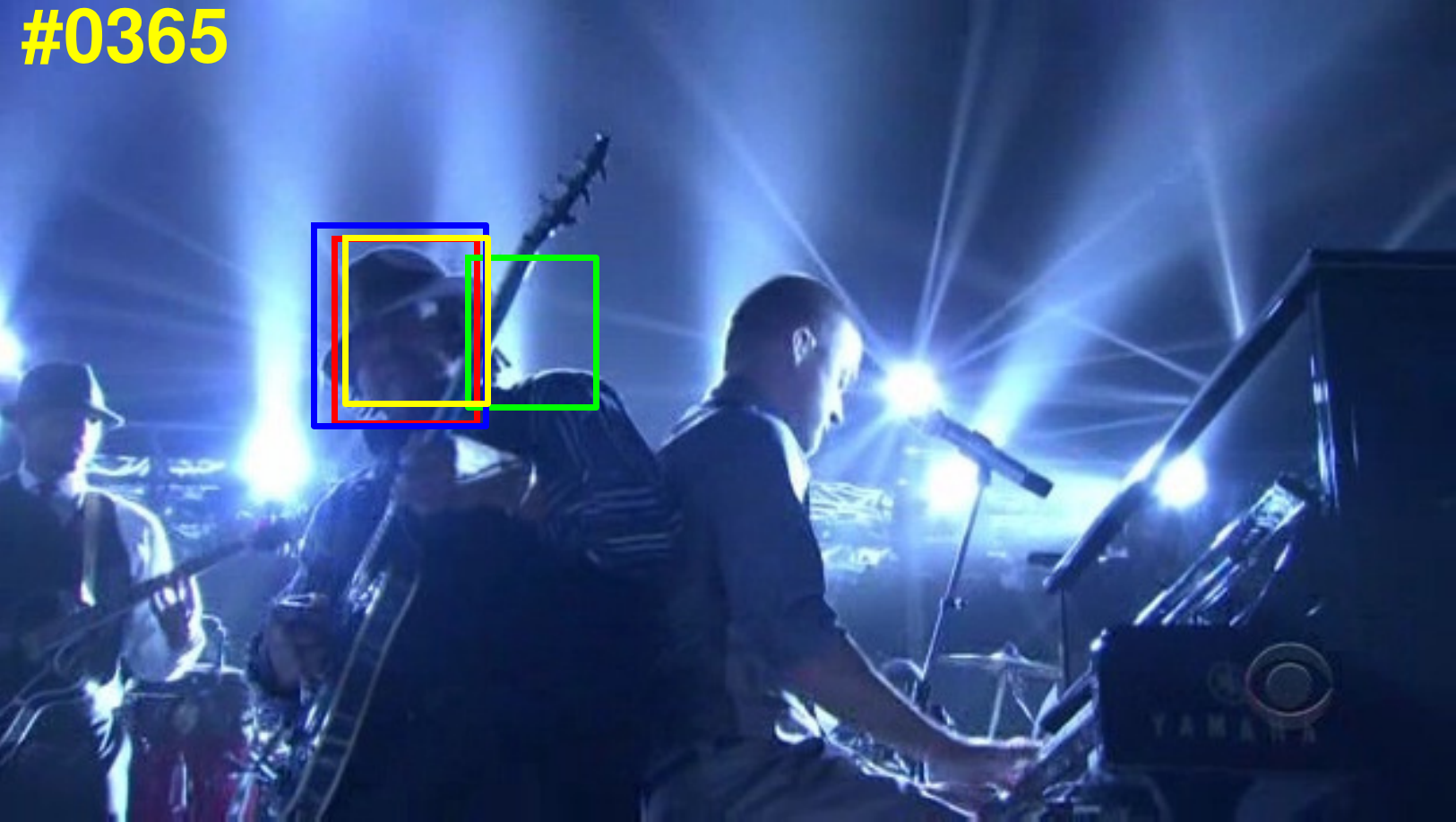}\\
\end{tabular}
\begin{tabular}{c}
\includegraphics[width=\swone]{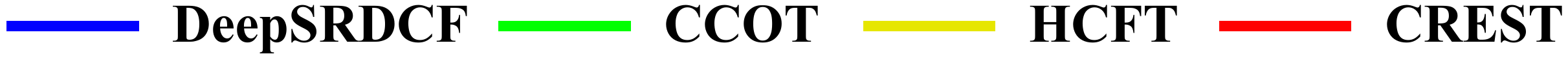}\\
\end{tabular}
\end{center}
\vspace{-5mm}
\caption{Convolutional features improve DCFs (DeepSRDCF \cite{Danelljan-iccvw15-DeepSRDCF}, CCOT \cite{martin-eccv16-beyond}, HCFT \cite{chao-iccv15-HCF}). We propose the CREST algorithm by formulating DCFs as a shallow convolutional layer with residual learning. \ryn{It} performs favorably against existing DCFs
with convolutional features.}
\label{fig:intro}
\end{figure}

However, existing DCFs based tracking algorithms are limited by two aspects.
First, learning DCFs is independent of feature extraction.
Although it is straightforward to learn DCFs directly over deep convolutional features as in \cite{chao-iccv15-HCF,Danelljan-iccvw15-DeepSRDCF,martin-eccv16-beyond}, DCFs trackers benefit little from the end-to-end training. Second, most DCFs trackers use a linear interpolation operation to update the learned filters over time. Such an empirical interpolation weight is unlikely to strike a good balance between model adaptivity and stability. It leads to \ryn{drifting of the} DCFs trackers due to noisy updates. \ryn{These limitations raise two questions: (1) whether DCFs with feature representation can be modeled in an end-to-end manner, and (2) whether DCFs can be updated in a more effective way rather than using those empirical operations such as linear interpolation?}

To address these two questions, we propose a Convolutional RESidual learning scheme for visual Tracking (CREST). We interpret DCFs as the counterparts of the convolution filters in deep neural networks. In light of this idea, we reformulate DCFs as a one-layer convolutional neural network that directly generates the response map as the spatial correlation between two consecutive frames. With this formulation, feature extraction through pre-trained CNN models (e.g., VGGNet \cite{simonyan-iclr14-very}), correlation response map generation, as well as model update are effectively integrated into an end-to-end form. The spatial convolutional operation functions similarly with the dot product between the circulant shifted inputs and the correlation filter. It removes the boundary effect in Fourier transform through directly convolving in the spatial domain. Moreover, the convolutional layer is fully differentiable. It allows updating \ryn{the} convolutional filters using back propagation.
Similar to DCFs, the convolutional layer generates dense response scores over all searching locations in a one-pass manner. To properly update our model, we apply residual learning \cite{he-cvpr16-resnet} to capture appearance changes by detecting the difference between the output of this convolutional layer and ground truth soft label. This helps alleviate a rapid model degradation caused by noisy updates. Meanwhile, residual learning contributes to the target response robustness for large appearance variations. Ablation studies (Section \ref{sec:ablation}) show that the proposed convolutional layer performs well against state-of-the-art DCFs trackers and the residual learning \ryn{approach} further improves the accuracy.

The main contributions of this work are as follows:
\begin{itemize}[noitemsep,nolistsep]
  \item We reformulate the correlation filter as one convolutional layer. It integrates feature extraction, response generation, and model update into the convolutional neural network for end-to-end training.
  \item We apply residual learning to capture the target appearance changes referring to spatiotemporal frames. This effectively alleviates a rapid model degradation by large appearance changes.
  \item We extensively validate our method on benchmark datasets with large-scale sequences. We show that our CREST tracker performs favorably against state-of-the-art trackers.
\end{itemize}

\section{Related Work}
There are extensive surveys of visual tracking in the literature \cite{yilmaz-acm06-object,salti-tip12-adaptive,smeulders-pami14-visual}. In this section, we mainly discuss tracking methods that are based on correlation filters and \ryn{CNNs}.

{\flushleft \bf Tracking by Correlation Filters.} Correlation filters for visual tracking have attracted considerable attention due to the computational efficiency in the Fourier domain. Tracking methods based on correlation filters regress all the circular-shifted versions of the input features to a Gaussian function. They do not need multiple samples of target appearance. The MOSSE tracker \cite{bolme-cvpr10-mosse} encodes target appearance through an adaptive correlation filter by optimizing the output sum of squared error. Several extensions have been proposed to considerably improves tracking accuracy. The examples include kernelized correlation filters \cite{Henriques-eccv12-DCF}, multiple dimensional features \cite{martin-cvpr14-adaptive,henriques-pami15-high}, context learning \cite{zhang-eccv14-fast}, scale estimation \cite{martin-bmvc14-accurate}, re-detection \cite{ma-cvpr15-lct}, subspace learning \cite{liu-cvpr15-real}, short-term and long-term memory \cite{hong-cvpr15-muster}, reliable collection \cite{li-cvpr15-reliable} and spatial regularization \cite{martin-iccv15-learning}. Different from existing correlation filters based frameworks that formulate correlation operation as an element wise multiplication in the Fourier domain, we formulate the correlation filter as a convolution operation in the spatial domain. It is presented by one convolutional layer in CNN. In this sense, we demonstrate that feature extraction, response generation as well as model update can be integrated into one network for end-to-end prediction and optimization.

\begin{figure*}[t]
\begin{center}
\begin{tabular}{c}
\includegraphics[width=0.9\linewidth]{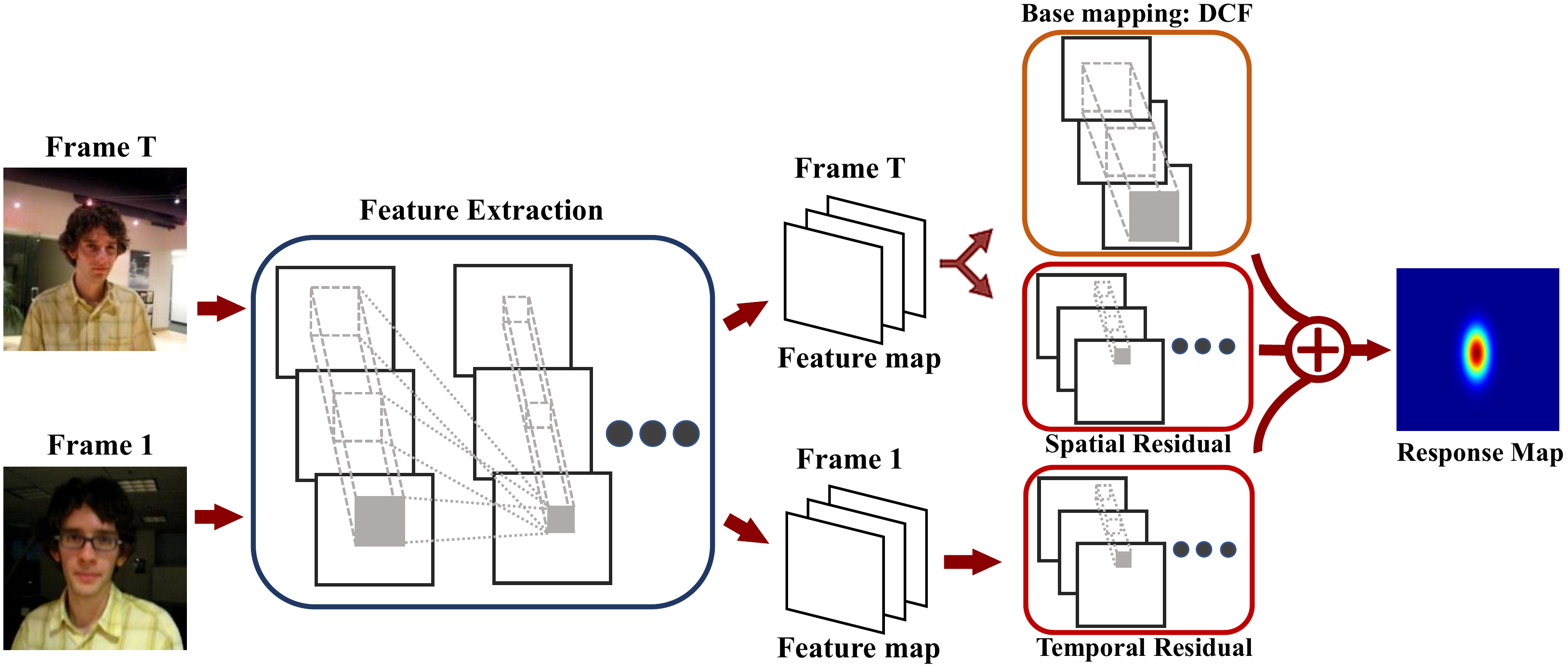}\\
\end{tabular}
\end{center}
\vspace{-5mm}
\caption{The pipeline of our CREST algorithm. We extract convolutional features from one search patch of the current frame T and the initial frame. These features are transformed into the response map through the base and residual \ryn{mappings}.}
\label{fig:pipeline}
\vspace{-3mm}
\end{figure*}

{\flushleft \bf Tracking by CNNs.}
Visual representations are important for visual tracking. Existing CNN trackers mainly explore the pre-trained object recognition networks and build upon discriminative or regression models. Discriminative tracking methods propose multiple particles and refine them through online classification. They include stacked denoising autoencoder \cite{wang-nips13-dlt}, incremental learning \cite{Li-bmvc14-deeptrack}, SVM classification \cite{hong-icml15-cnnsvm} and fully connected neural network \cite{nam-cvpr16-mdnet}. These discriminative trackers require auxiliary training data as well as an off-line pre-training. On the other hand, regression based methods typically regress CNN features into soft labels (e.g., a two dimensional Gaussian distribution). They  focus on integrating convolutional features with the traditional DCF framework. The examples include hierarchical convolutional features \cite{chao-iccv15-HCF}, adaptive hedging \cite{qi-cvpr16-hdt}, spatial regularization \cite{Danelljan-iccvw15-DeepSRDCF} and continuous convolutional operations \cite{martin-eccv16-beyond}. In addition, there are methods based on CNN to select convolutional features \cite{wang-iccv15-visual} and update sequentially \cite{wang-2016-stct}. Furthermore, the Siamese networks receive growing attention due to its two stream identical structure. These include tracking by object verification \cite{tao-cvpr16-siamese}, tracking by correlation \cite{bertinetto-eccv16-fully} and tracking by location axis prediction \cite{held-eccv16-learning}. Besides, there is investigation on the recurrent neural network (RNN) to facilitate tracking as object verification \cite{cui-cvpr16-rnn}. Different from the existing frameworks, we apply residual learning to capture the difference of the predicted response map between the current frame and the ground-truth (the initial frame). This facilitates to account for appearance changes and effectively reduce model degradation caused by noisy updates.

\section{Convolutional Residual Learning}
Our CREST algorithm carries out feature regression through the base and residual layers. The base layer consists of one convolutional layer which is formulated as the traditional DCFs. The difference between the base layer output and ground truth soft label is captured through the residual layers. Figure \ref{fig:pipeline} shows the \ryn{CREST pipeline. The details are discussed below.}

\subsection{DCF Reformulation}\label{sec:base}
We revisit the DCF based framework and formulate it as our base layer.
The DCFs learn a discriminative classifier and predict the target translation through searching the maximum value in the response map. We denote the input sample by $X$ and denote the corresponding Gaussian function label by $Y$. A correlation filter $W$ is then learned by solving the following minimization problem:
\begin{equation}
W^\star=\mathop{\arg\min}\limits_{W}||W\ast X-Y||^2+\lambda||W||^2,
\label{eq:dcf}
\end{equation}
where $\lambda$ is the regularization parameter. Typically, the convolution operation between the correlation filter $W$ and the input $X$ is formulated into a dot product in the Fourier domain \cite{Henriques-eccv12-DCF,ma-cvpr15-lct,henriques-pami15-high}.

We reformulate the learning process of DCFs as the loss minimization of the convolutional neural network. The general form of the loss function \cite{jia-arxiv14-caffe} can be written as:
\begin{equation}
L(W)=\frac{1}{N}\sum_i^{|N|}\mathcal{L}_W(X^{(i)})+\lambda r(W),
\label{eq:loss}
\end{equation}
where $N$ is the number of samples, $\mathcal{L}_W(X^{(i)})$ ($i\in N$) is the loss of the $i$-th sample, and $r(W)$ is the weight decay. We set $N=1$ and take the L2 norm as $r(W)$. The loss function in Eq. \ref{eq:loss} can be written as:
\begin{equation}
L(W)=\mathcal{L}_W(X)+\lambda||W||^2,
\label{eq:lossL2}
\end{equation}
where $\mathcal{L}_W(X)=||\mathcal{F}(X)-Y||^2$. It is equivalent to the L2 loss between $\mathcal{F}(X)$ and $Y$ where $\mathcal{F}(X)$ is the network output and $Y$ is the ground truth label. We take $\mathcal{F}(X)=W\ast X$ as the convolution operation on $X$, which can be achieved through one convolutional layer. The convolutional filters $W$ is equivalent to the correlation filter and the loss function in Eq. \ref{eq:lossL2} is equivalent to the DCFs objective function. As a result, we formulate the DCFs as one convolutional layer with L2 loss as the objective function. It is named as the base layer in our network. Its filter size is set to cover the target object. The convolutional weights can be effectively calculated using the gradient descent method instead of the closed form solution \cite{henriques-pami15-high}.

\subsection{Residual Learning}\label{sec:res}
We formulate DCFs as a base layer represented by one convolutional layer. Ideally, the response map from the base layer output will be identical to the ground truth soft label. In practice, it is unlikely that a single layer network is able to accomplish that. Instead of stacking more layers which may cause the degradation problem \cite{he-cvpr16-resnet}, we apply the residual learning to effectively capture the difference between the base layer output and the ground truth.

\begin{figure}[t]
\begin{center}
\begin{tabular}{c}
\includegraphics[width=1.0\linewidth]{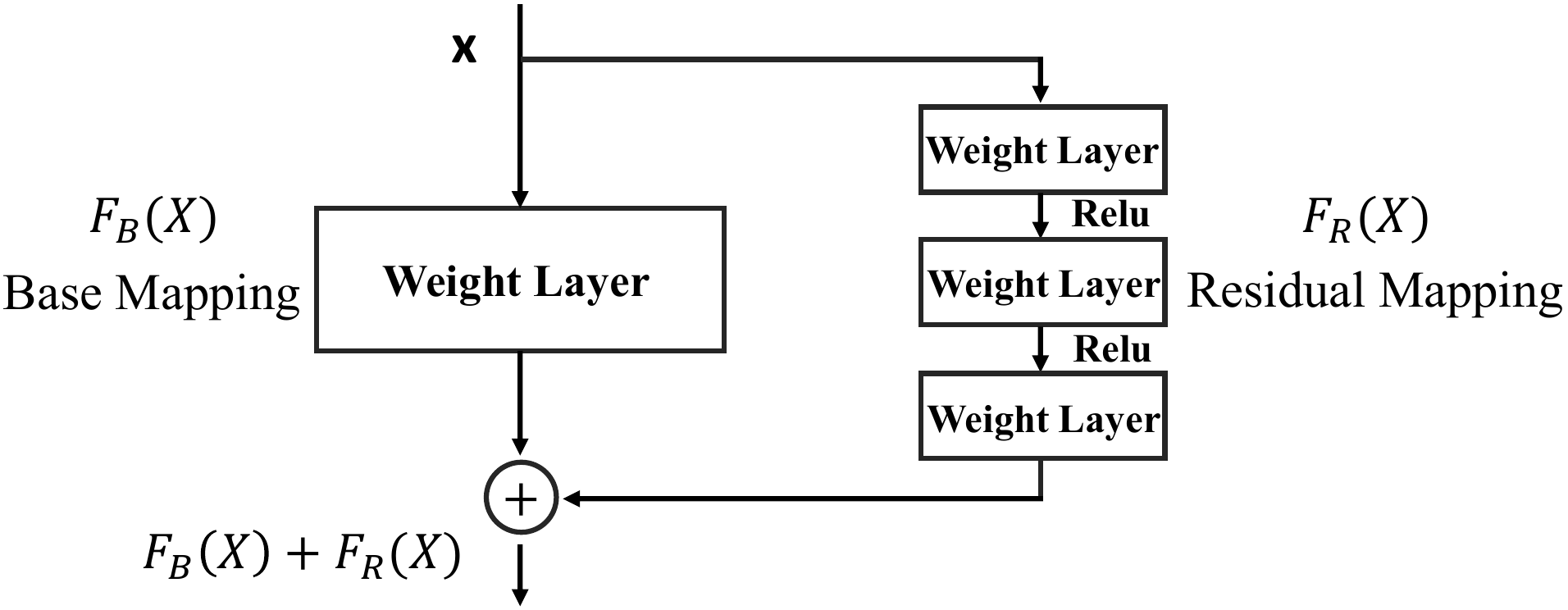}\\
\end{tabular}
\end{center}
\vspace{-3mm}
\caption{Base and spatial residual layers.}
\label{fig:res}
\end{figure}

Figure \ref{fig:res} shows the structure of the base and residual layers. We denote $\mathcal{H}(X)$ as the optimal mapping of input $X$ and $\mathcal{F}_B(X)$ as the output from the base layer. Rather than stacking more layers to approximate $\mathcal{H}(X)$, we expect these layers to approximate the residual function: $\mathcal{F}_{R}(X)=\mathcal{H}(X)-\mathcal{F}_B(X)$. As a result, our expected network output can be formulated as follows:
\begin{eqnarray}\label{eq:res}
\mathcal{F}(X)&=&\mathcal{F}_B(X)+\mathcal{F}_{R}(X)\nonumber\\
&=&\mathcal{F}_B(X,\{W_B\})+\mathcal{F}_{R}(X,\{W_{R}\}),
\end{eqnarray}
where $\mathcal{F}_B(X)=W_B\ast X$. The mapping $\mathcal{F}_{R}(X,\{W_{R}\})$ represents the residual learning and $W_{R}$ is a general form of convolutional layers with biases and ReLU \cite{nair-icml10-relu} omitted for simplifying notations. We adopt three layers in the residual learning with small filter size. They are set to capture the residual which is not presented in the base layer output. Finally, input $X$ is regressed through the base and residual mapping to generate the output response map.

In addition, we can also utilize the temporal residual which helps to capture the difference when the spatial residual is not effective. We develop a temporal residual learning network that contains similar structure as the spatial residual learning. The temporal input is extracted from the first frame which contains the initial object appearance. Let $X_t$ denote the input $X$ on frame $t$.
Thus we have
\begin{equation} \mathcal{F}(X_\textrm{t})=\mathcal{F}_R(X_\textrm{t})+\mathcal{F}_{SR}(X_\textrm{t})+\mathcal{F}_{TR}(X_\textrm{1}),\\
\label{eq:rest}
\end{equation}
where $\mathcal{F}_{TR}(X_\textrm{1})$ is the temporal residual from the first frame. The proposed spatiotemporal residual learning process encodes elusive object representation into the response map generation framework and no additional data is needed to train the network.

Figure \ref{fig:algo_visual} shows one example of the filter response from the base and residual layers. The feature maps are scaled for visualization purpose. Given an input image, we first crop the search patch centered at the estimated position of the previous frame. The patch is sent into our feature extraction network and then regressed into a response map through the base and residual layers. We observe that the base layer performs similarly to the traditional DCF based trackers to predict response map. When target objects undergo small appearance variations, the difference between the base layer output and the ground truth is minor. The residual layers have little effect on the final response map. However, when target objects undergo large appearance variations, such as background clutter, the response from the base layer is limited and may not differentiate the target and the background. Nevertheless, this limitation is alleviated by the residual layers, which effectively model the difference between the base layer output and the ground truth. It helps to reduce the noisy response values on the final output through the addition of base and residual layers. As a result, target response is more robust to large appearance variations.

\def\swone{1.0\linewidth}
\begin{figure}[t]
\begin{center}
\begin{tabular}{c}
\includegraphics[width=\swone]{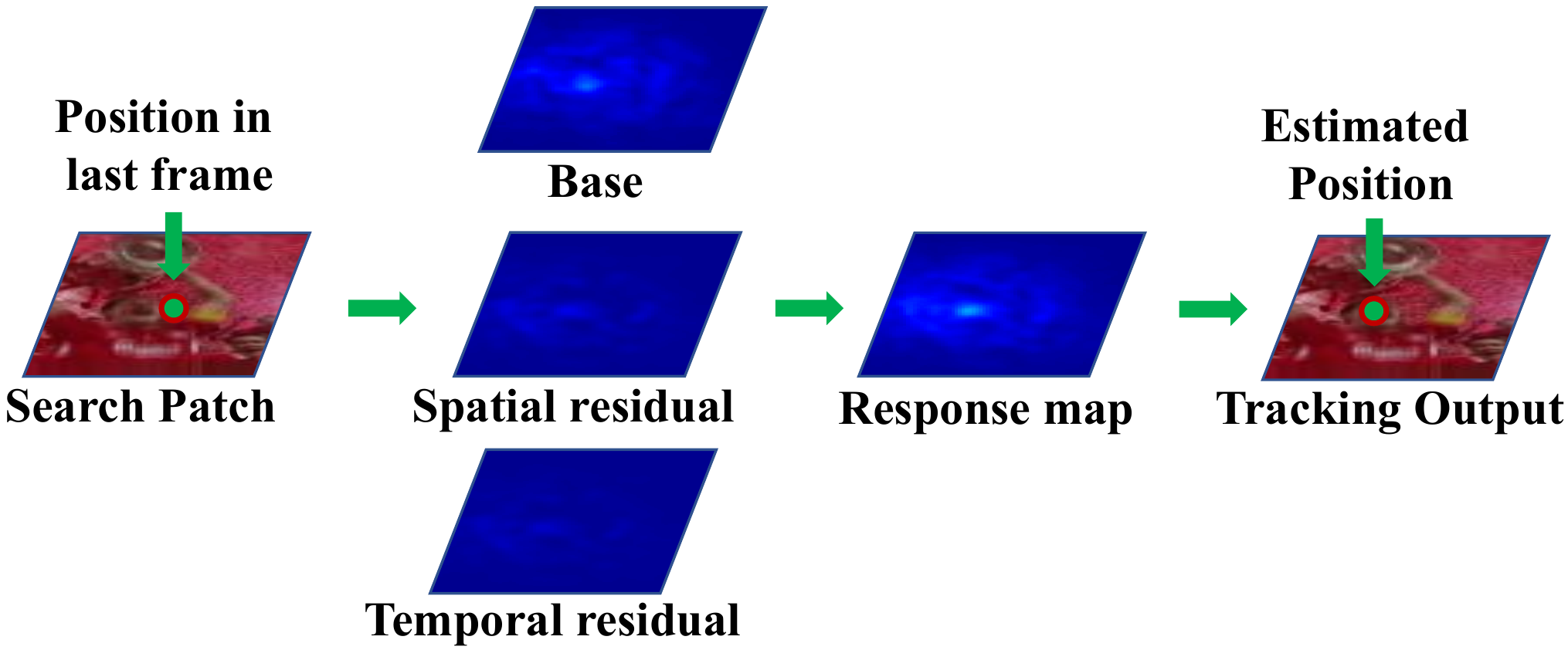}\\
\end{tabular}
\end{center}
\vspace{-5mm}
\caption{Visualization of \ryn{the} feature maps in the target localization step.}
\label{fig:algo_visual}
\vspace{-3mm}
\end{figure}

\section{Tracking via CREST}
We illustrate the detailed procedure of CREST for visual tracking.
As we do not need offline training we present the tracking process through the aspects of model initialization, online detection, scale estimation and model update.

{\flushleft \bf Model Initialization.}
Given an input frame with the target location, we extract a training patch which is centered on the target object. This patch is sent into our framework for feature extraction and response mapping. We adopt the VGG network \cite{simonyan-iclr14-very} for feature extraction. Meanwhile, all the parameters in the base and residual layers are randomly initialized following zero mean Gaussian distribution. Our base and residual layers are well initialized after a few steps.

{\flushleft \bf Online Detection.}
The online detection scheme is straightforward. When a new frame comes we extract the search patch from the center location predicted by the previous frame. The search patch has the same size with the training patch and fed into our framework to generate a response map. Once we have the response map, we locate the object by searching for the maximum response value.

{\flushleft \bf Scale Estimation.}
After we obtain the target center location, we extract search patches in different scales. These patches are then resized into a fixed size of training patches.
Thus the candidate object sizes in different scales are all normalized. We send these patches into our network to generate the response map. The width $w_\textrm{t}$ and height $h_\textrm{t}$ of the target object at frame $t$ is updated as:
\begin{equation}
(w_\textrm{t},h_\textrm{t})=\beta(w_\textrm{t}^\star,h_\textrm{t}^\star)+(1-\beta)(w_{\textrm{t}-1},h_{\textrm{t}-1}),
\label{eq:scale}
\end{equation}
where $w_\textrm{t}^\star$ and $h_\textrm{t}^\star$ are the width and height of the scaled object with maximum response value. The weight factor $\beta$ enables the smooth update of the target size.

{\flushleft \bf Model Update.}
We consistently generate training data during online tracking. For each frame, after predicting the target location we can generate the corresponding ground truth response map, and the search patch can be directly adopted as the training patch. The collected training patches and response maps from every $T$ frames are adopted as training pairs which will be fed into our network for online update.

\section{Experiments}
In this section, we introduce the implementation details and analyze the effect of each component including the base and residual layers. We then compare our CREST tracker with state-of-the-art trackers on the benchmark datasets for performance evaluation.

\subsection{Experimental Setups}

{\flushleft \bf Implementation Details.}
We obtain the training patch from the first frame. It is 5 times the maximum value of object width and height. Our feature extraction network is from VGG-16 \cite{simonyan-iclr14-very} with only the first two pooling layers retained. We extract the feature maps from the \textit{conv4-3} layer and reduce the feature channels to 64 through PCA dimensionality reduction, which is learned using the first frame image patch. The regression target map is generated using a two-dimensional Gaussian function with a peak value of 1.0. The weight factor $\beta$ for scale estimation is set to 0.6. Our experiments are performed on a PC with an i7 3.4GHz CPU and a GeForce GTX Titan Black GPU with MatConvNet toolbox \cite{Vedaldi-mm15-matconvnet}. In the training stage, we iteratively apply the adam optimizer \cite{Diederik-iclr15-adam} with a learning rate of 5e-8 to update the coefficients, until the loss in Eq. \ref{eq:lossL2} is below the given threshold of 0.02. We observe that in practice, our network converges from random initialization in a few hundred iterations. We update the model every 2 frames for only 2 iterations with a learning rate of 2e-9.

{\flushleft \bf Benchmark Datasets.}
The experiments are conducted on three standard benchmarks: OTB-2013 \cite{wu-cvpr13-otb}, OTB-2015 \cite{wu-pami15-otb} and VOT-2016 \cite{kristan-eccvw16-vot}. The first two datasets contain 50 and 100 sequences, respectively. They are annotated with ground truth bounding boxes and various attributes. In the VOT-2016 dataset, there are 60 challenging videos from a set of more than 300 videos.

{\flushleft \bf Evaluation Metrics.}
We follow the standard evaluation metrics from the benchmarks. For the OTB-2013 and OTB-2015 we use the one-pass evaluation (OPE) with precision and success plots metrics. The precision metric measures the rate of frame locations within a certain threshold distance from those of the ground truth. The threshold distance is set as 20 for all the trackers. The success plot metric measures the overlap ratio between predicted and ground truth bounding boxes. For the VOT-2016 dataset, the performance is measured in terms of expected average overlap (EAO), average overlap (AO), accuracy values (Av), accuracy ranks (Ar), robustness values (Rv) and robustness ranks (Rr). The average overlap is similar to the AUC metric in OTB benchmarks.

\def\swtwo{0.495\linewidth}
\renewcommand{\tabcolsep}{.1pt}
\begin{figure}[t]
\begin{center}
\begin{tabular}{cc}
\includegraphics[width=\swtwo]{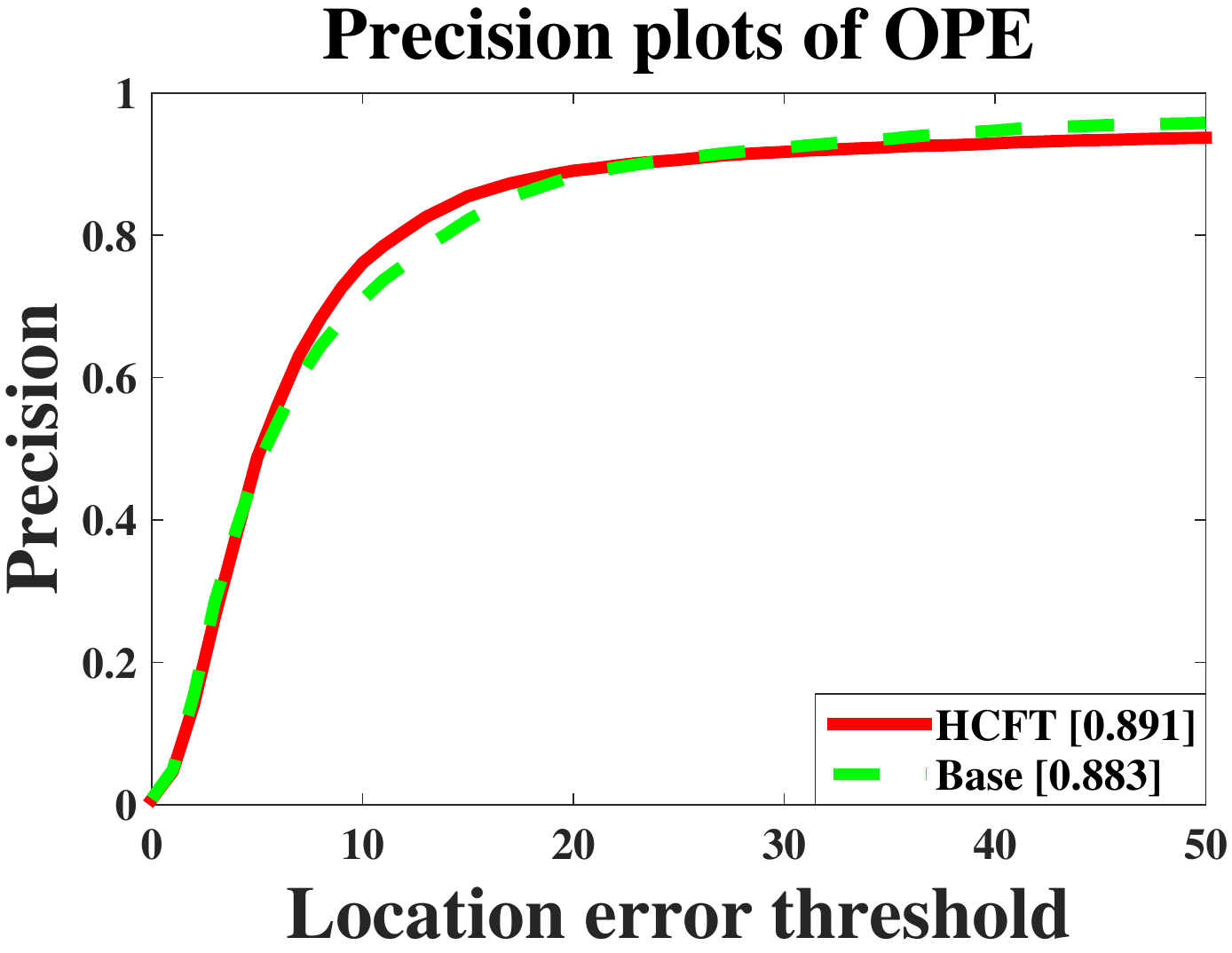}&
\includegraphics[width=\swtwo]{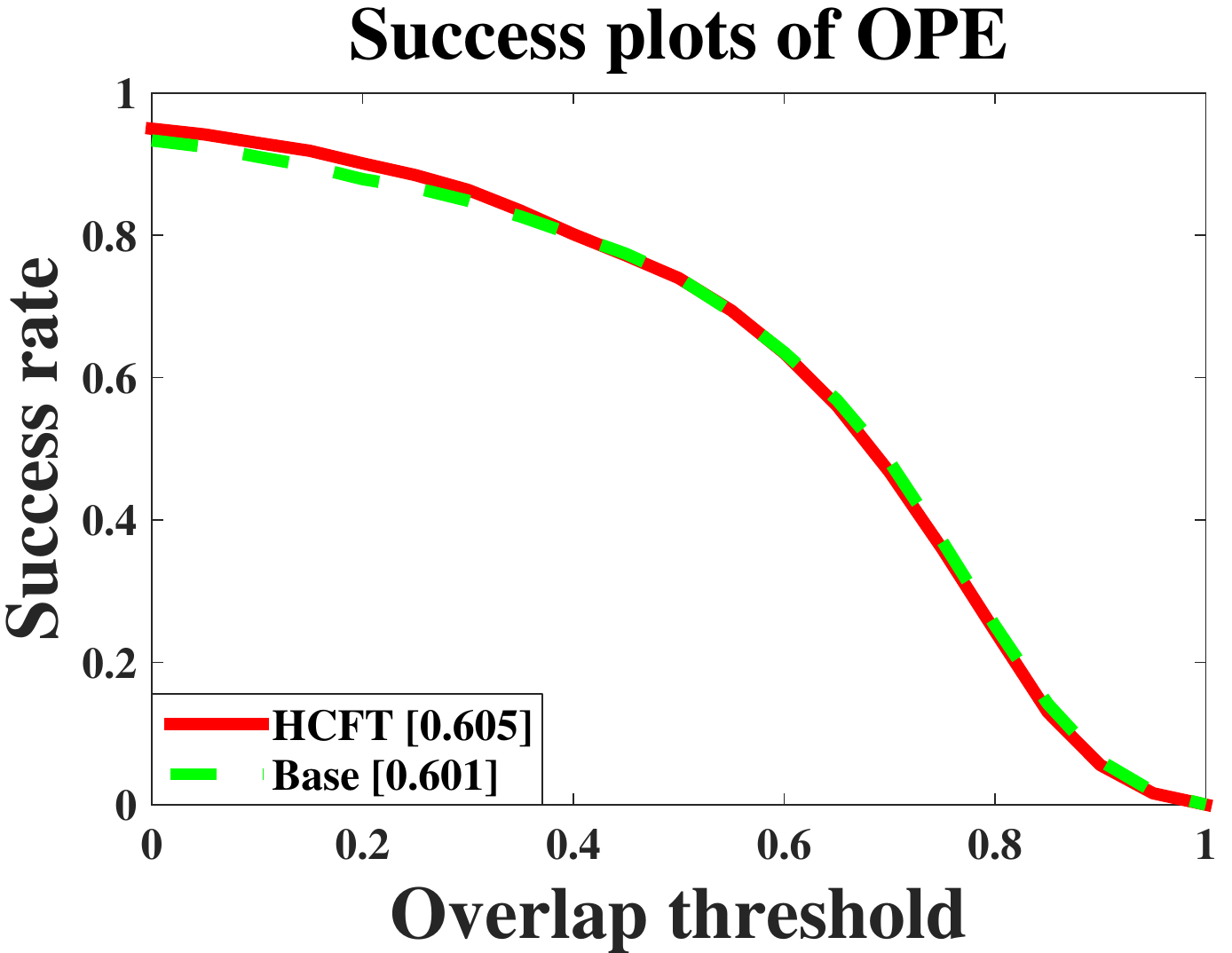}
\end{tabular}
\end{center}
\vspace{-5mm}
\caption{Precision and success plots using one-pass evaluation on the OTB-2013 dataset.
The performance of the base layer without scale estimation is similar \ryn{to} that of HCFT \cite{chao-iccv15-HCF} on average.}
\label{fig:abla1}
\begin{center}
\begin{tabular}{cc}
\includegraphics[width=\swtwo]{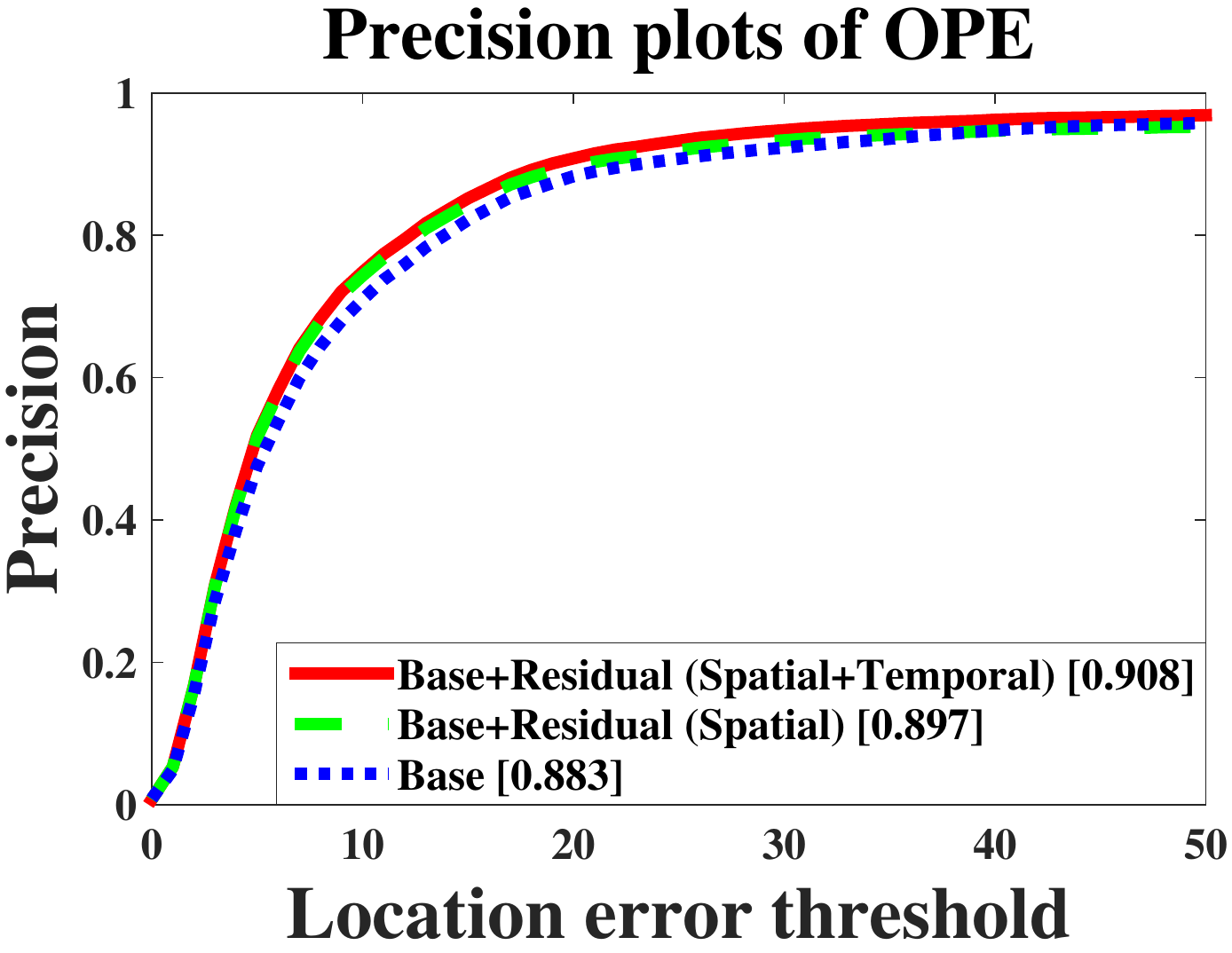}&
\includegraphics[width=\swtwo]{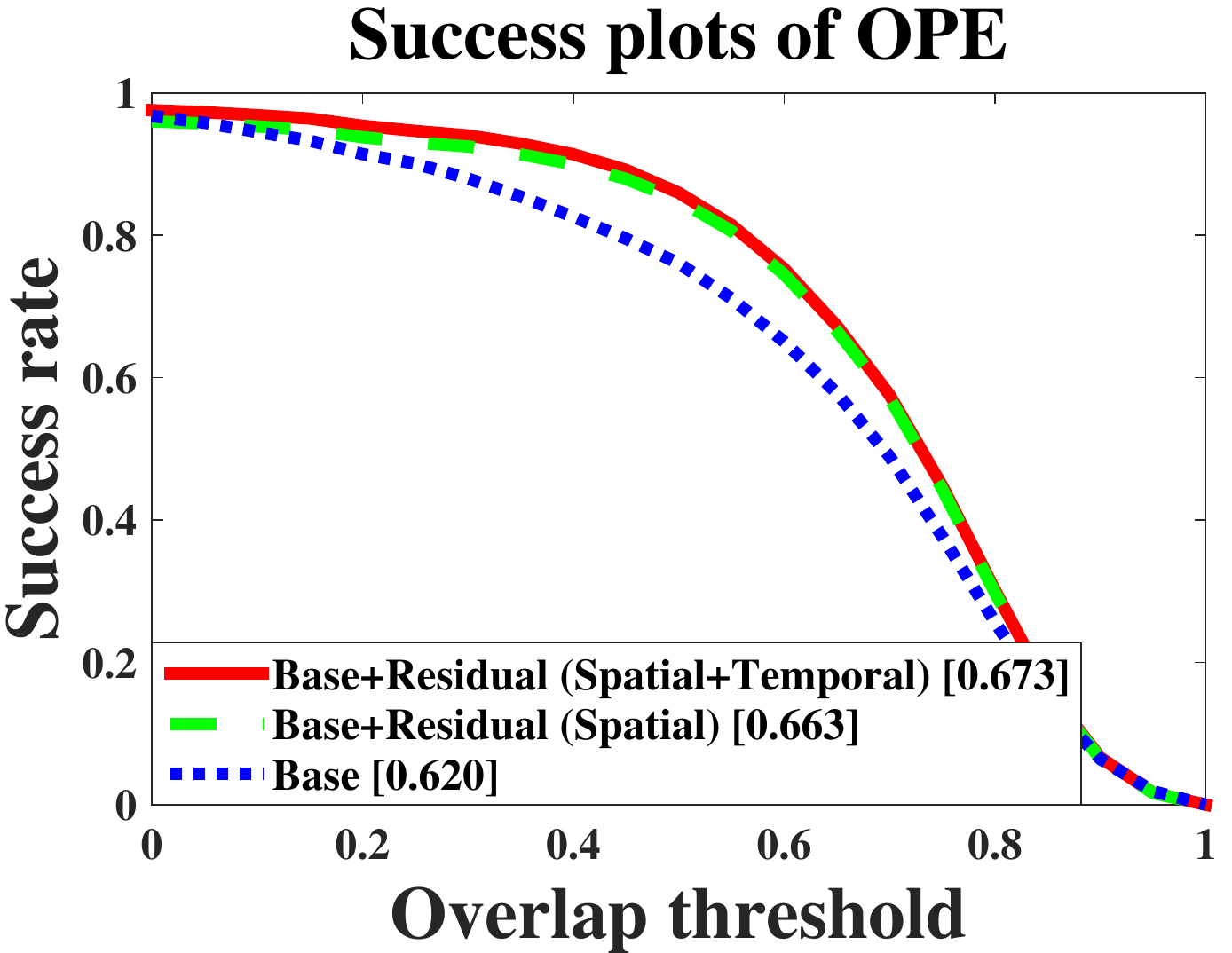}
\end{tabular}
\end{center}
\vspace{-5mm}
\caption{Precision and success plots using one-pass evaluation on the OTB-2013 dataset.
The performance of the base layer is improved gradually with the integration of spatiotemporal residual.}
\label{fig:abla2}
\vspace{-3mm}
\end{figure}

\subsection{Ablation Studies}\label{sec:ablation}

Our CREST algorithm contains base and residual layers. As analyzed in Section \ref{sec:base}, the base layer is formulated similar to existing DCF based trackers, and the residual layers refine the response map. In this section, we conduct ablation analysis to compare the performance of the base layer with those of the DCF based trackers. In addition, we also evaluate the base layer and its integration with spatiotemporal residual layers.

We analyze our CREST algorithm in the OTB-2013 dataset. We first compare the base layer performance with that of HCFT \cite{chao-iccv15-HCF}, which is a traditional DCF based tracker with convolutional features. The objective function of HCFT is shown in Eq. \ref{eq:dcf}, which is the same as ours. As there is no scale estimation in HCFT, we remove this step in our algorithm. Figure \ref{fig:abla1} shows the quantitative evaluation under AUC and average distance precision scores. We observe that the performance of our base layer is similar to that of HCFT on average. It indicates that our base layer achieves similar performance as the DCF based trackers with convolutional features. The DeepSRDCF \cite{Danelljan-iccvw15-DeepSRDCF} and CCOT \cite{martin-eccv16-beyond} trackers are different from the traditional DCF based trackers because they add a spatial constraint on the regularization term, which is different from our objective function. We also analyze the effect of residual integration in Figure \ref{fig:abla2}. The AUC and average distance precision scores show that the base layer obtains obvious improvement through the integration of spatial residual learning. Meanwhile, temporal residual contributes little to the overall performance.

\subsection{Comparisons with State-of-the-art}

\def\swtwo{0.49\linewidth}
\renewcommand{\tabcolsep}{.1pt}
\begin{figure}[t]
\begin{center}
\begin{tabular}{cc}
\includegraphics[width=\swtwo]{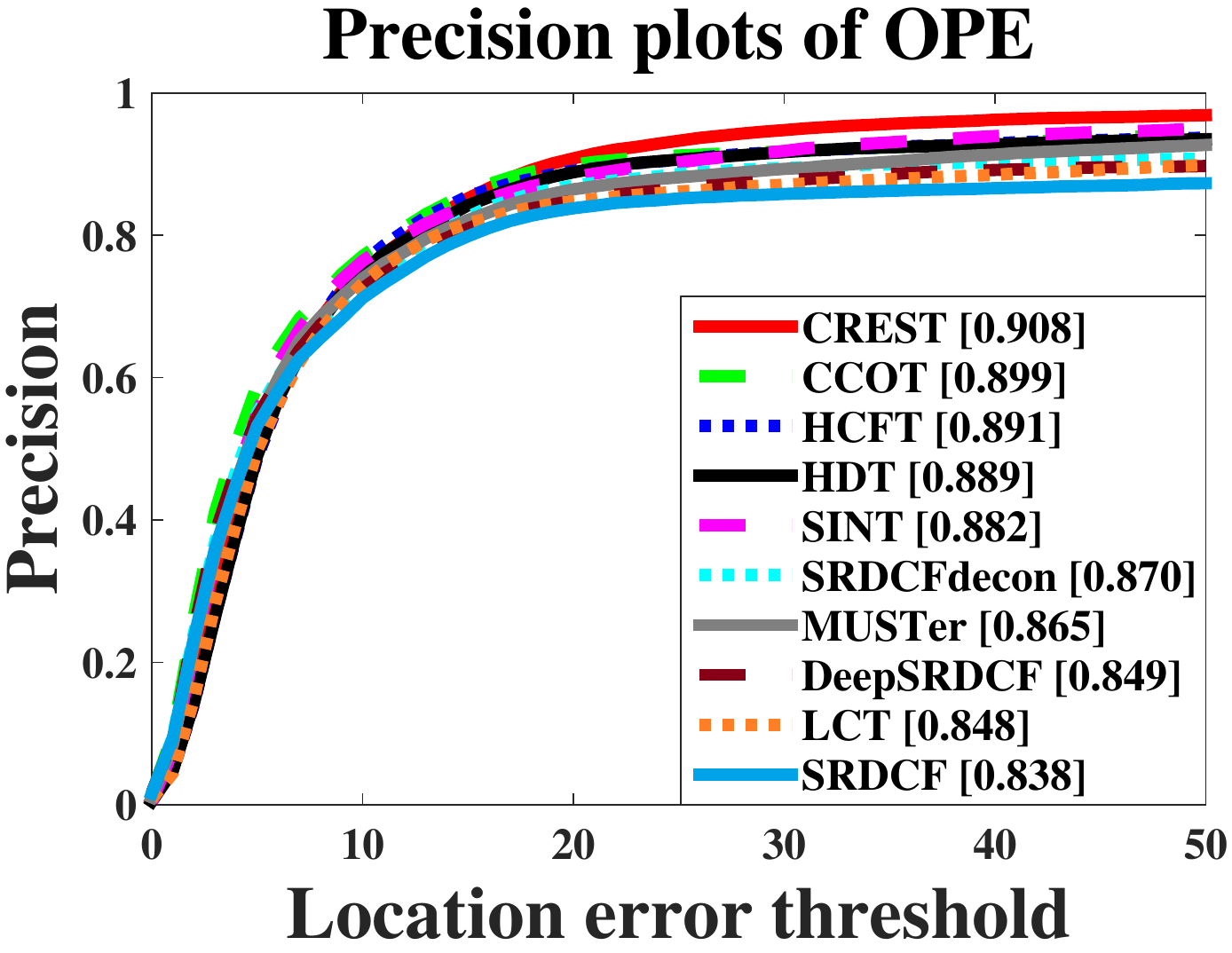}&
\includegraphics[width=\swtwo]{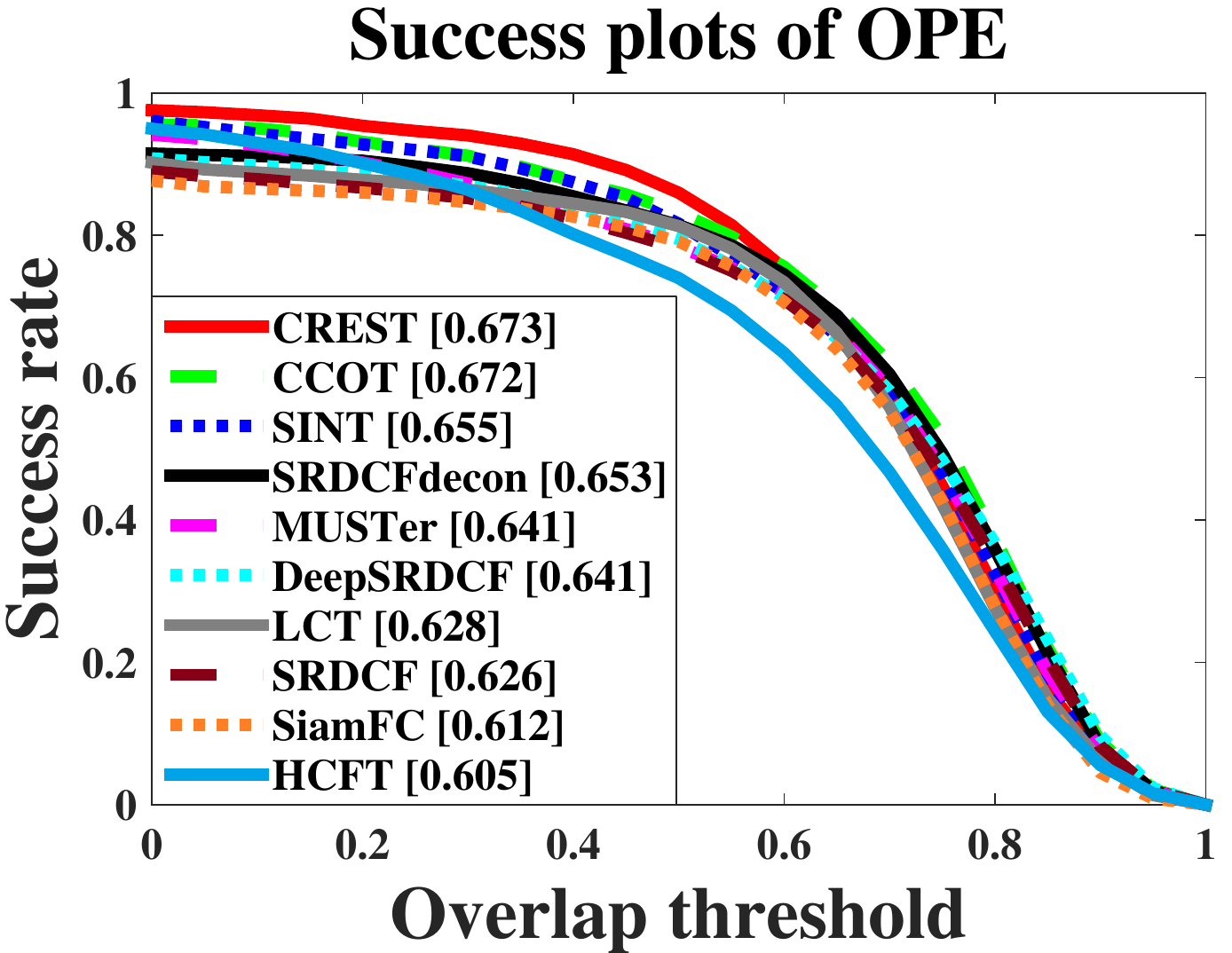}
\end{tabular}
\end{center}
\vspace{-5mm}
\caption{Precision and success plots over all 50 sequences using one-pass evaluation on the OTB-2013 Dataset. The legend contains the area-under-the-curve score and the average distance precision score at 20 pixels for each tracker.}
\label{fig:otb2013}
\end{figure}

We conduct quantitative and qualitative evaluations on the benchmark datasets including OTB-2013 \cite{wu-cvpr13-otb}, OTB-2015 \cite{wu-pami15-otb} and VOT-2016 \cite{kristan-eccvw16-vot}. The details are discussed \ryn{in the following}.

\subsubsection{Quantitative Evaluation}

{\flushleft \bf OTB-2013 Dataset.}
We compare our CREST tracker with 29 trackers from the OTB-2013 benchmark \cite{wu-cvpr13-otb} and other 21 state-of-the-art trackers including KCF \cite{Henriques-eccv12-DCF}, MEEM \cite{zhang-eccv14-meem}, TGPR \cite{gao-eccv14-transfer}, DSST \cite{martin-bmvc14-accurate}, RPT \cite{li-cvpr15-reliable}, MUSTer \cite{hong-cvpr15-muster}, LCT \cite{ma-cvpr15-lct}, HCFT \cite{chao-iccv15-HCF}, FCNT \cite{wang-iccv15-visual}, SRDCF \cite{martin-iccv15-learning}, CNN-SVM \cite{hong-icml15-cnnsvm}, DeepSRDCF \cite{Danelljan-iccvw15-DeepSRDCF}, DAT \cite{possegger-cvpr15-defense}, Staple \cite{bertinetto-cvpr16-staple}, SRDCFdecon \cite{danelljan-CVPR16-adaptive}, CCOT \cite{martin-eccv16-beyond}, GOTURN \cite{held-eccv16-learning}, SINT \cite{tao-cvpr16-siamese}, SiamFC \cite{bertinetto-eccv16-fully}, HDT \cite{qi-cvpr16-hdt} and SCT \cite{choi-cvpr16-visual}. The evaluation is conducted on 50 video sequences using one-pass evaluation with distance precision and overlap success metrics.

\def\swtwo{0.49\linewidth}
\renewcommand{\tabcolsep}{.1pt}
\begin{figure}[t]
\begin{center}
\begin{tabular}{cc}
\includegraphics[width=\swtwo]{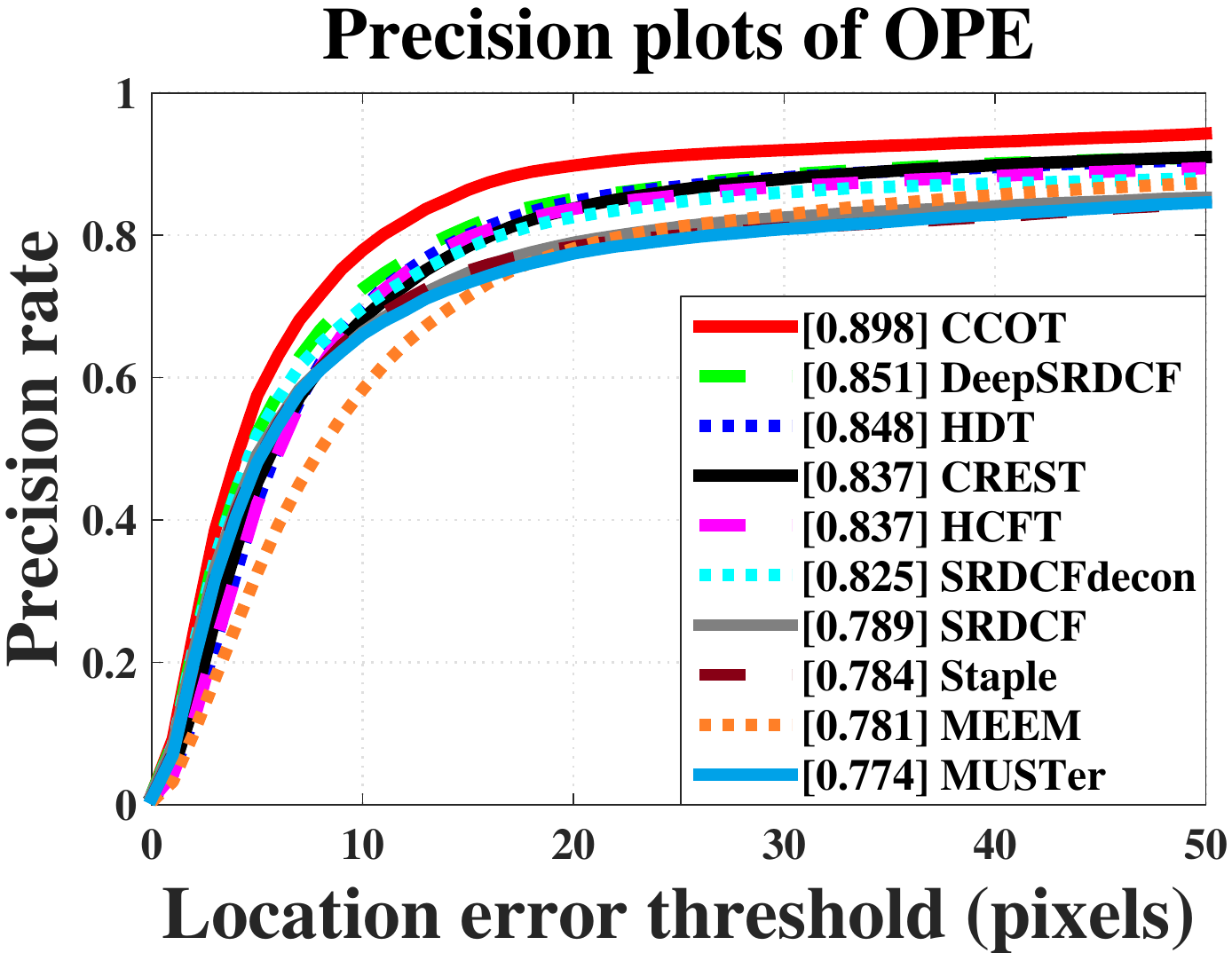}&
\includegraphics[width=\swtwo]{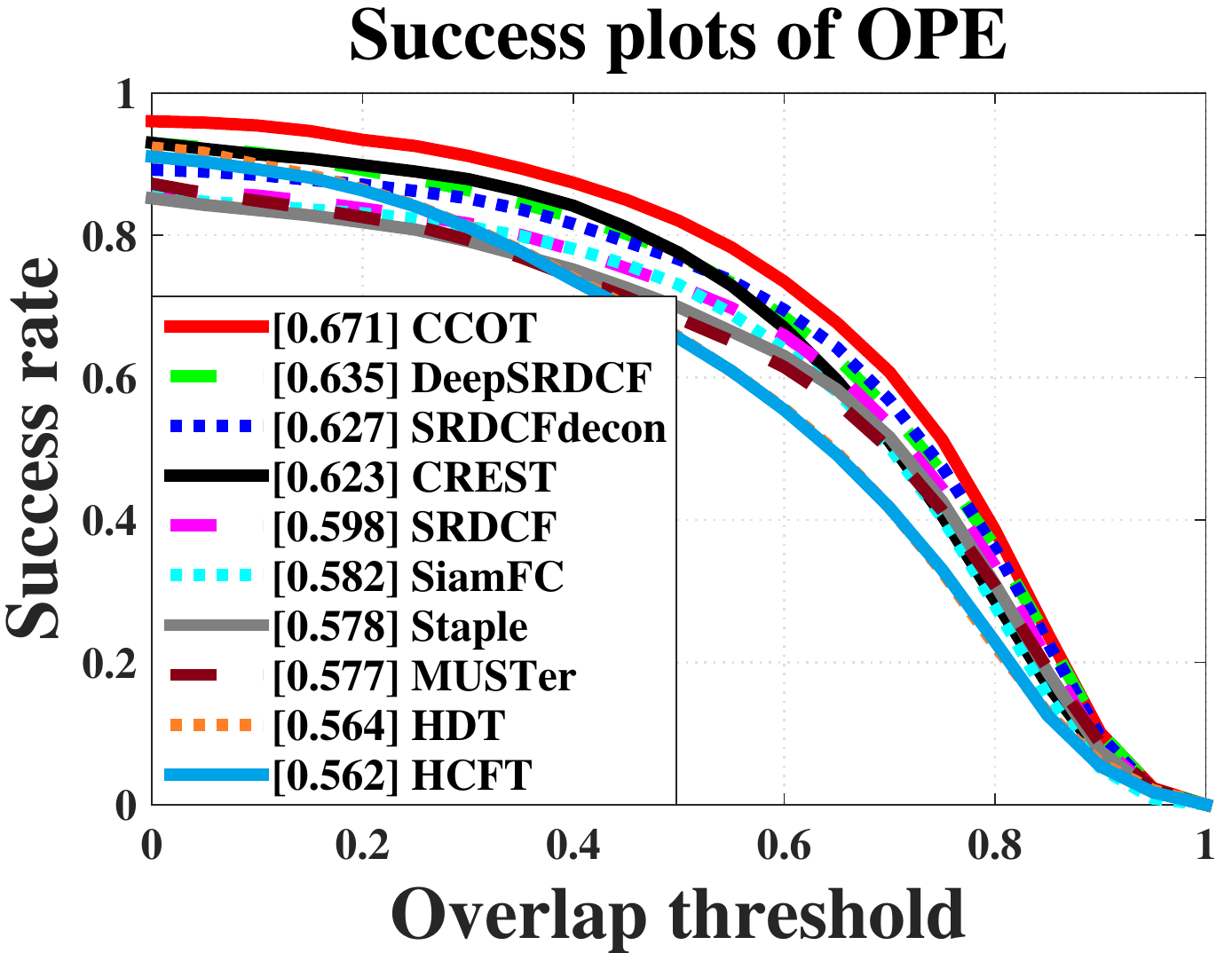}
\end{tabular}
\end{center}
\vspace{-5mm}
\caption{Precision and success plots over all 100 sequences using one-pass evaluation on the OTB-2015 Dataset. The legend contains the area-under-the-curve score and the average distance precision score at 20 pixels for each tracker.}
\label{fig:otb2015}
\end{figure}

Figure \ref{fig:otb2013} shows the evaluation results. We only show the top 10 trackers for presentation clarity. The AUC and distance precision scores for each tracker are reported in the figure legend. Among all the trackers, our CREST tracker performs favorably on both the distance precision and overlap success rate. In Figure \ref{fig:otb2013}, we exclude the MDNet tracker \cite{nam-cvpr16-mdnet} as it uses tracking videos for training. Overall, the precision and success plots demonstrate that our CREST tracker performs favorably against state-of-the-art trackers.

\def\swfour{0.245\linewidth}
\renewcommand{\tabcolsep}{.1pt}
\begin{figure*}[t]
\begin{center}
\begin{tabular}{cccc}
\includegraphics[width=\swfour]{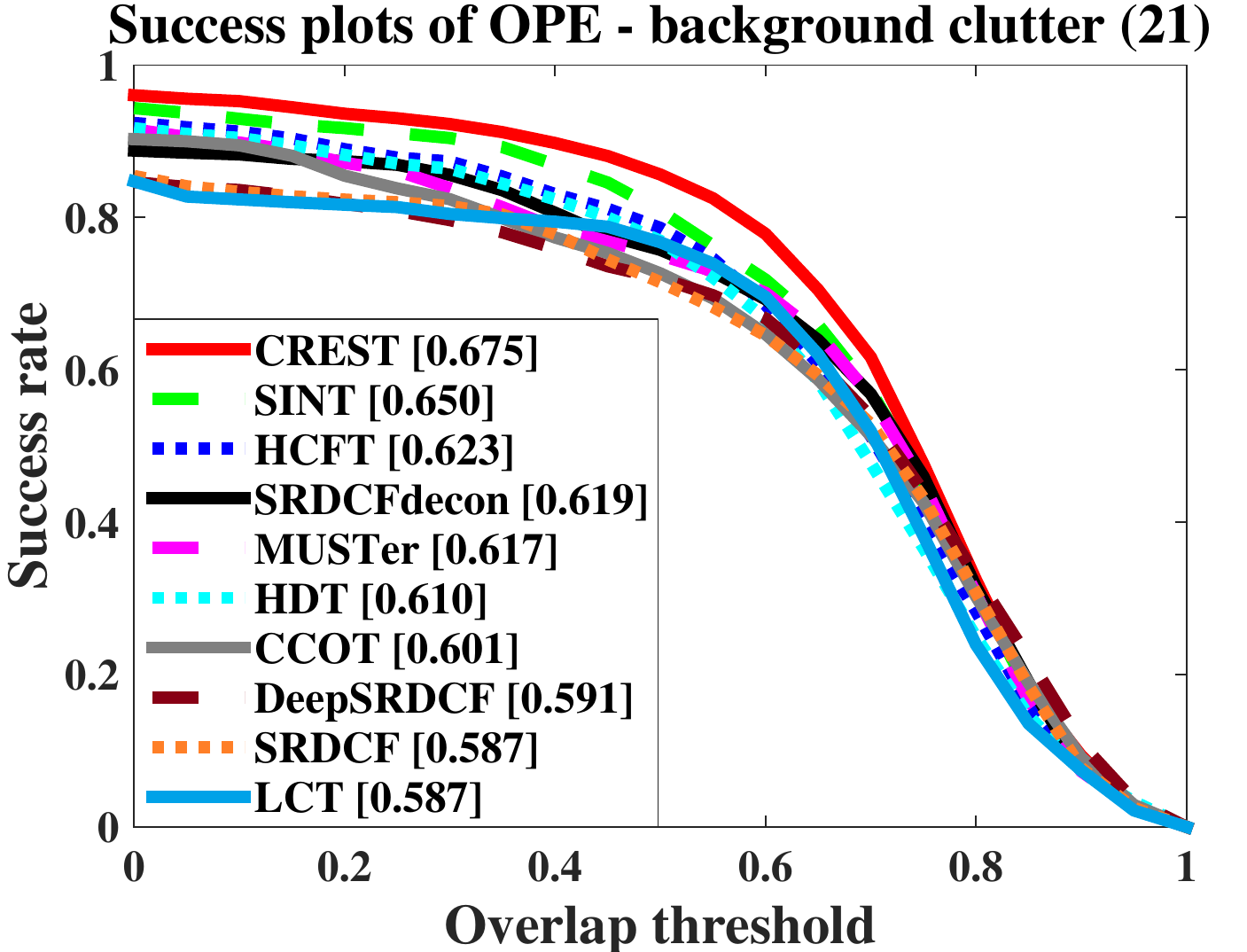}&
\includegraphics[width=\swfour]{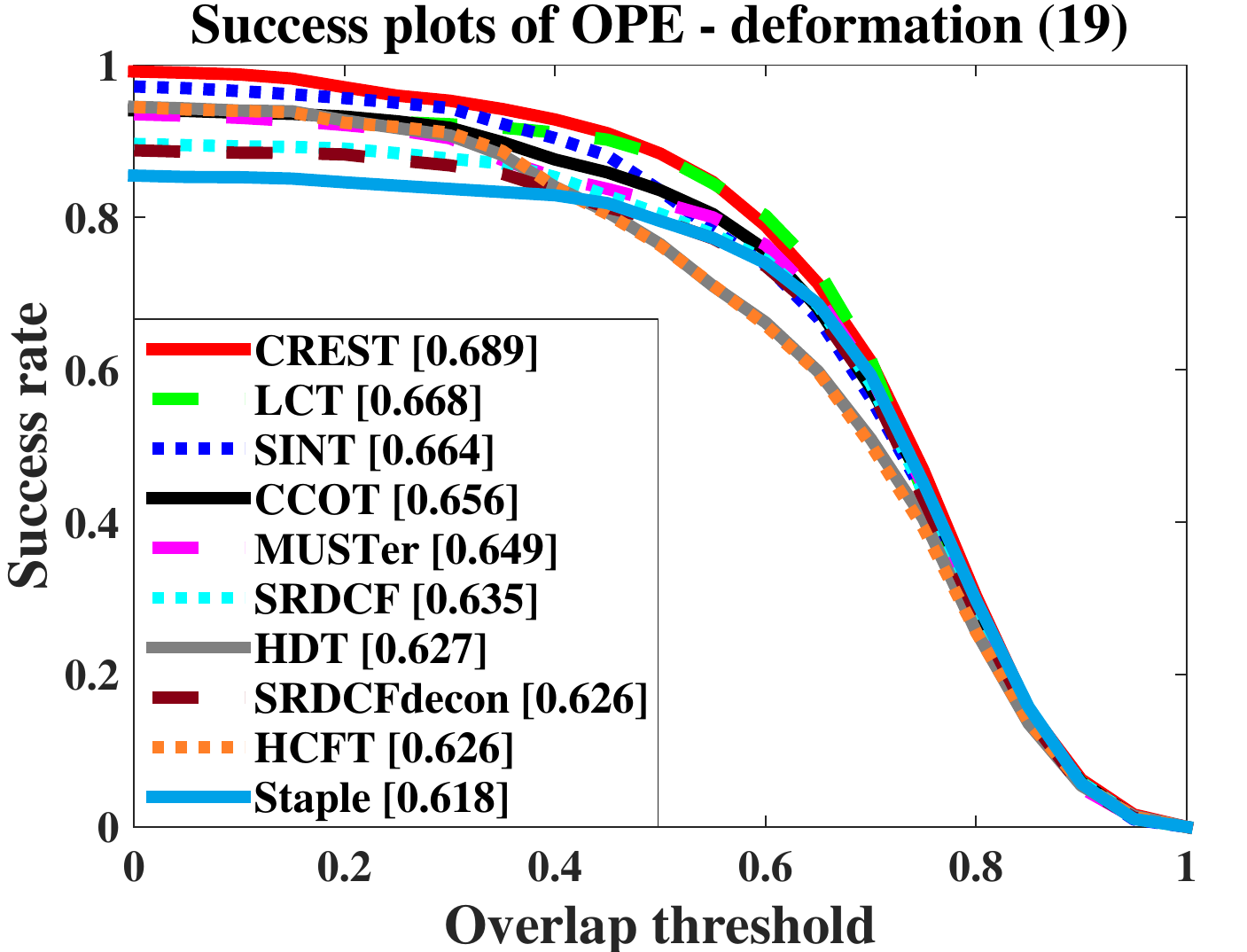}&
\includegraphics[width=\swfour]{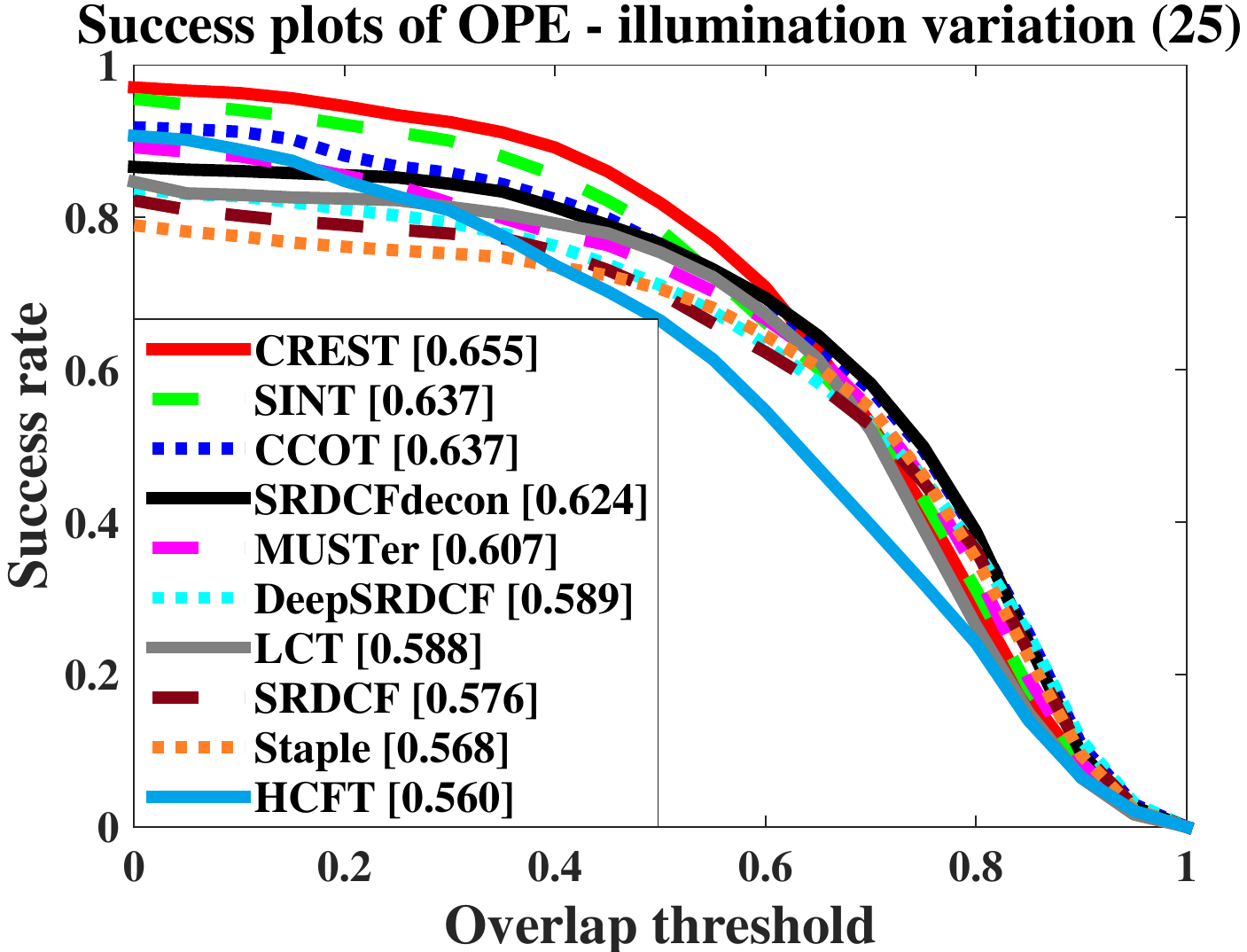}&
\includegraphics[width=\swfour]{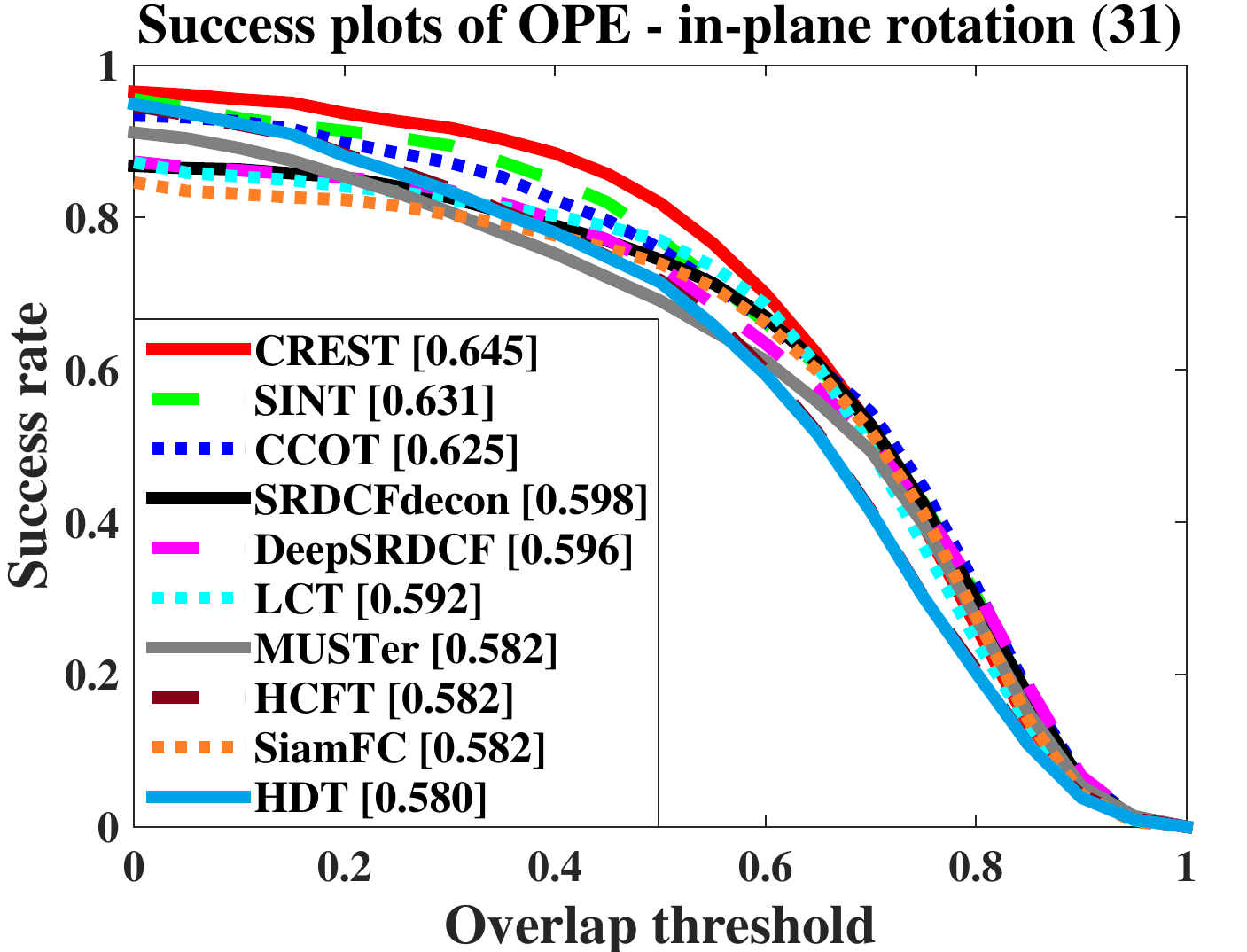}\\
\includegraphics[width=\swfour]{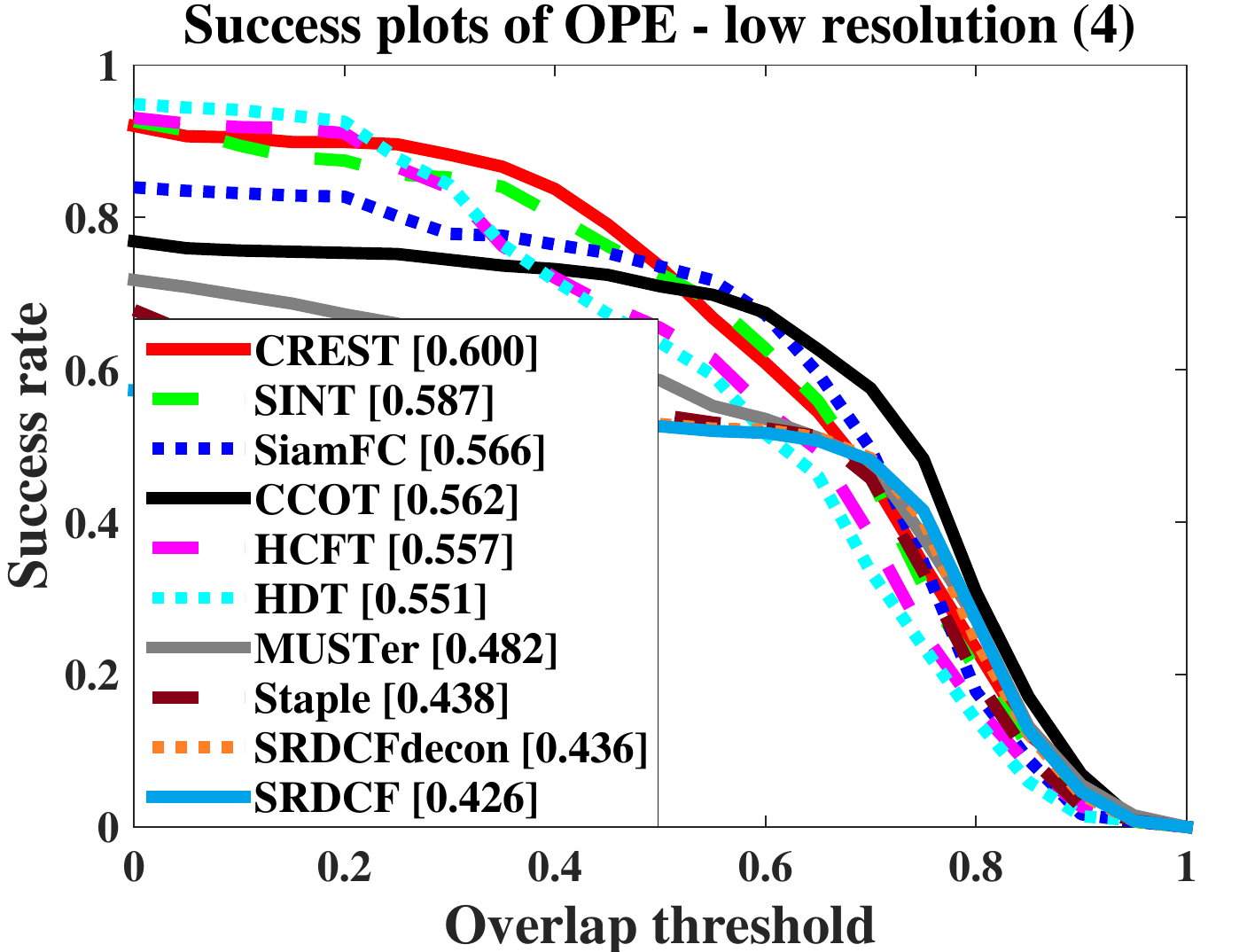}&
\includegraphics[width=\swfour]{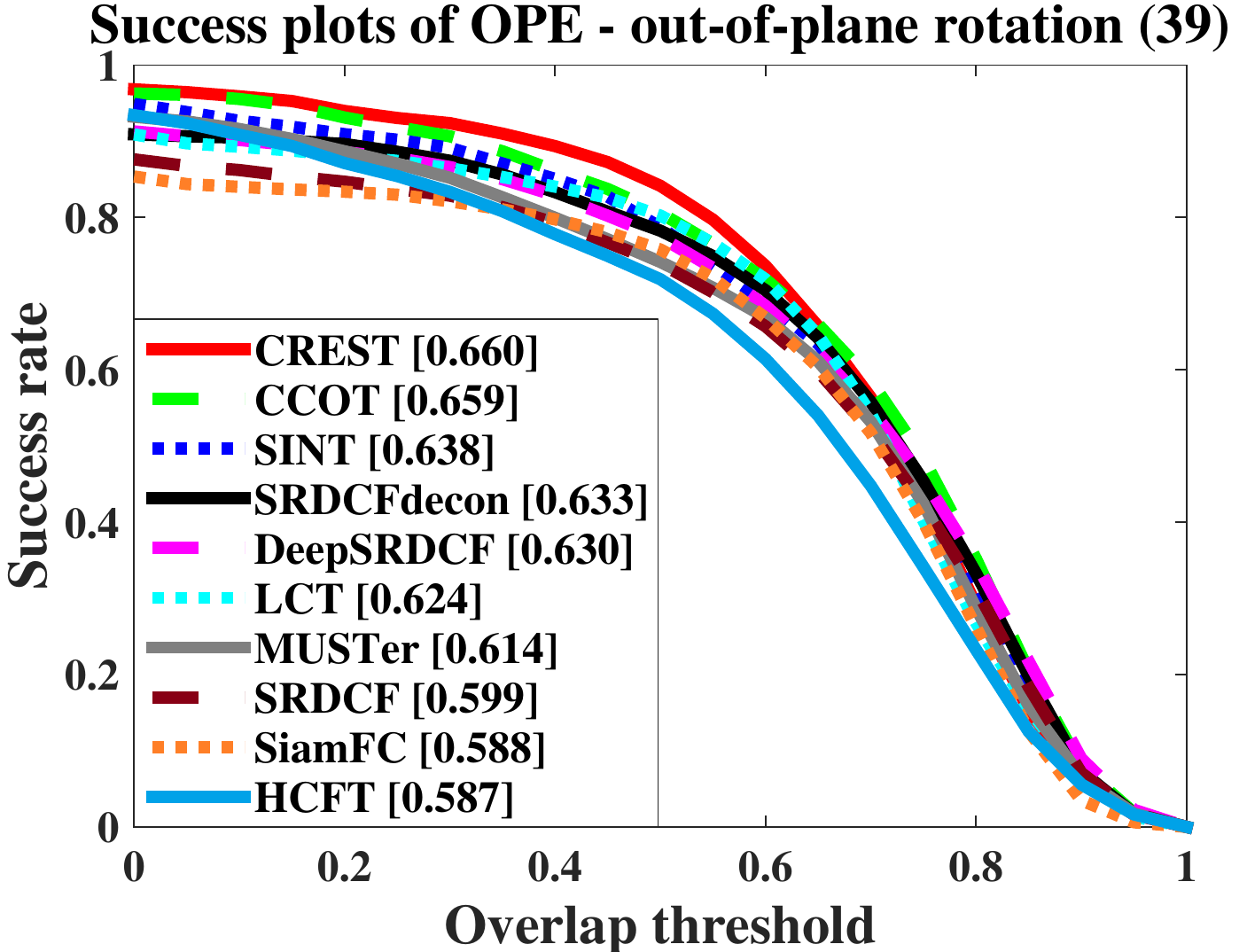}&
\includegraphics[width=\swfour]{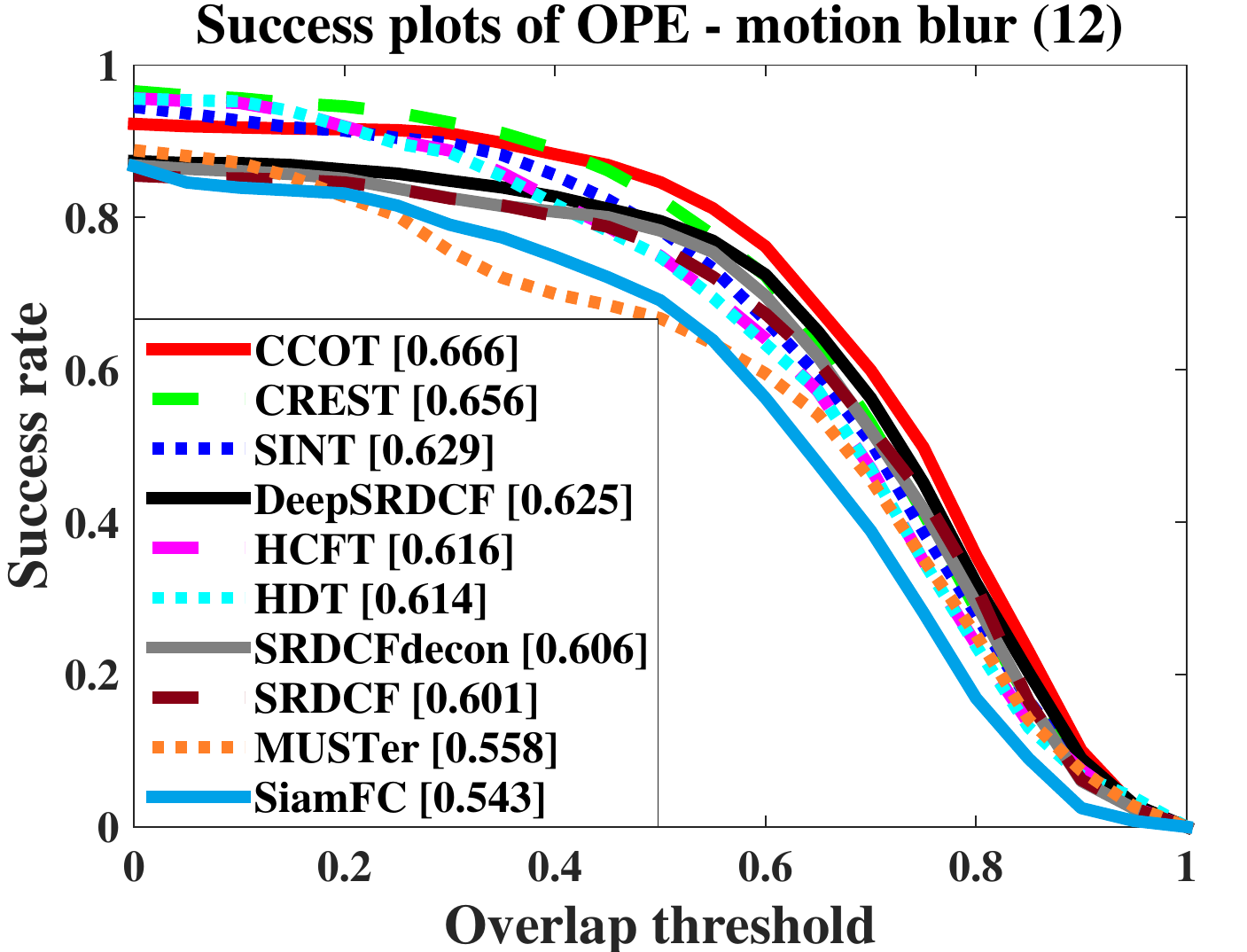}&
\includegraphics[width=\swfour]{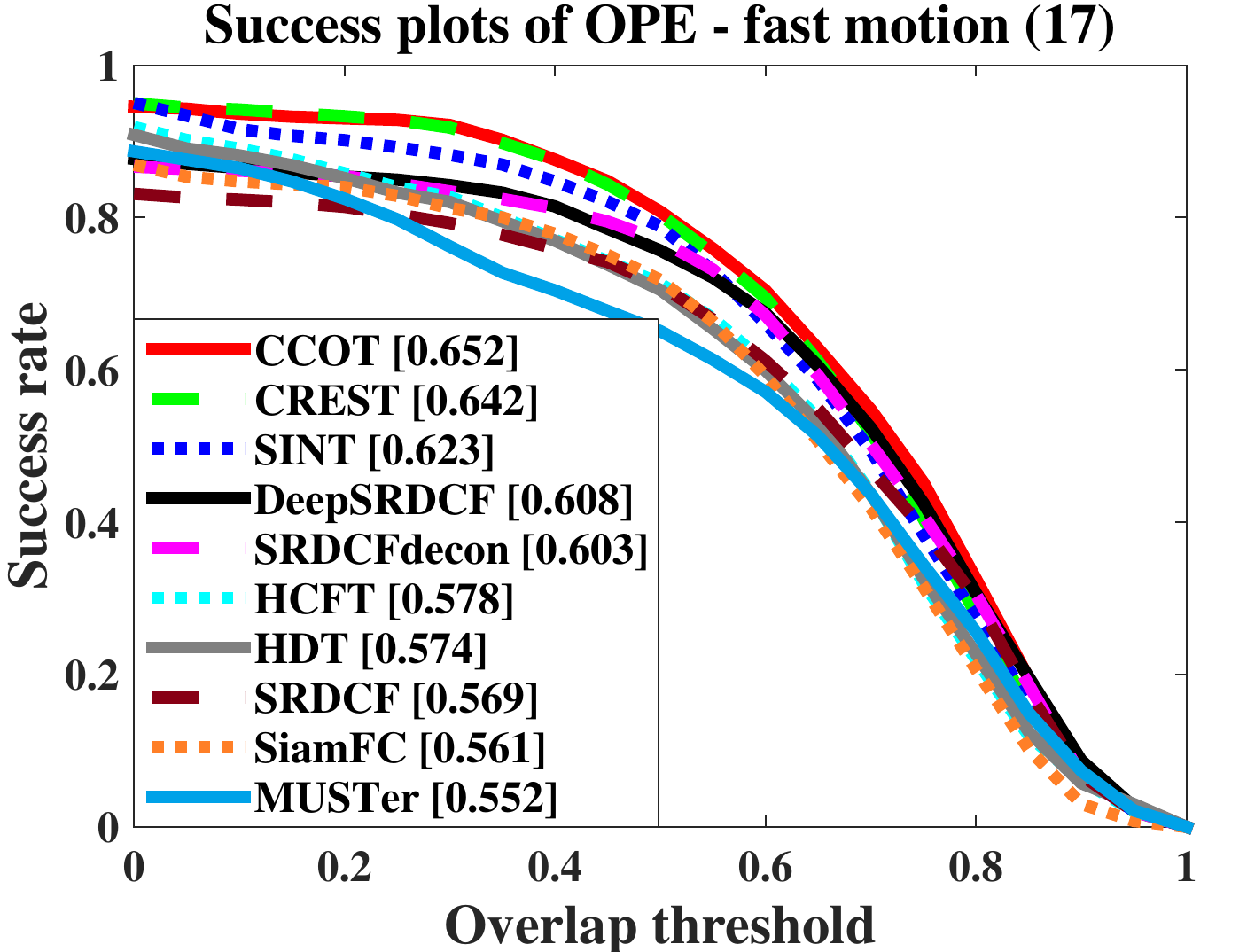}\\
\end{tabular}
\end{center}
\vspace{-5mm}
\caption{The success plots over eight tracking challenges, \ryn{including} background clutter, deformation, illumination variation, in-plan rotation, low resolution, out-of-plane rotation, motion blur and fast motion.}
\label{fig:attr}
\end{figure*}

In Figure \ref{fig:attr}, we further analyze the tracker performance under different video attributes (e.g., background clutter, deformation and illumination variation) annotated in the benchmark. We show the one pass evaluation on the AUC score under eight main video attributes. The results indicate that our CREST tracker is effective in handling background clutters and illumination variation. It is mainly because the residual layer can capture the appearance changes for effective model update. When the target appearance undergoes obvious changes or becomes similar to the background, existing DCF based trackers (e.g., HCFT, CCOT) can not accurately locate the target object. They do not perform well compared with SINT that makes sparse prediction through generating candidate bounding boxes and verifying through the Siamese network. This limitation is reduced in our CREST tracker, where the residual from the spatial and temporal domains effectively narrow the gap between the noisy output of the base layer and ground truth label. The dense prediction becomes accurate and the target location can be correctly identified. We have found similar performance in deformation, where LCT achieves a favorable result through integrating the redetection scheme. However, it does not perform well as our CREST tracker with the temporal residual integrated into the framework and optimized as a whole. In motion blur and fast motion sequences, our tracker does not perform well as CCOT. This can be attributed to the convolutional features from multiple layers CCOT has adopted. Their feature representation performs better than ours from single layer output. The feature representation from multiple layers will be considered in our future work.

\def\pp{\hspace{0mm}}
\renewcommand{\tabcolsep}{15pt}
\begin{table*}[t]
\caption{Comparison with \ryn{the} state-of-the-art trackers on the VOT 2016 dataset. The results are presented in terms of expected average overlap (EAO), average overlap (AO), accuracy value (Av), accuracy rank (Ar), robustness value (Rv), and robustness rank (Rr).}
\centering
       \begin{tabular}{cccccccc}
        \toprule
        &\small{CCOT}&\small{Staple}&\small{EBT}&\small{DSRDCF}&\small{MDNet}&\small{SiamFC}&\small{CREST}\\
        \midrule
        \small{EAO}&\small{0.331}&\small{0.295}&\small{0.291}&\small{0.276}&\small{0.257}&\small{0.277}&\small{0.283}\\
        \small{AO}&\small{0.469}&\small{0.388}&\small{0.370}&\small{0.427}&\small{0.457}&\small{0.421}&\small{0.435}\\
        \small{Av}&\small{0.523}&\small{0.538}&\small{0.441}&\small{0.513}&\small{0.533}&\small{0.549}&\small{0.514}\\
        \small{Ar}&\small{2.167}&\small{2.100}&\small{4.383}&\small{2.517}&\small{1.967}&\small{2.465}&\small{2.833}\\
        \small{Rv}&\small{0.850}&\small{1.350}&\small{0.900}&\small{1.167}&\small{1.204}&\small{1.382}&\small{1.083}\\
        \small{Rr}&\small{2.333}&\small{3.933}&\small{2.467}&\small{3.550}&\small{3.250}&\small{2.853}&\small{2.733}\\
        \bottomrule
       \end{tabular}
\label{tab:vot}
\end{table*}

{\flushleft \bf OTB-2015 Dataset.}
We also compare our CREST tracker on the OTB-2015 benchmark \cite{wu-pami15-otb} with the 29 trackers in \cite{wu-cvpr13-otb} and the state-of-the-art trackers, including KCF \cite{Henriques-eccv12-DCF}, MEEM \cite{zhang-eccv14-meem}, TGPR \cite{gao-eccv14-transfer}, DSST \cite{martin-bmvc14-accurate}, MUSTer \cite{hong-cvpr15-muster}, LCT \cite{ma-cvpr15-lct}, HCFT \cite{chao-iccv15-HCF}, SRDCF \cite{martin-iccv15-learning}, CNN-SVM \cite{hong-icml15-cnnsvm}, DeepSRDCF \cite{Danelljan-iccvw15-DeepSRDCF}, Staple \cite{bertinetto-cvpr16-staple}, SRDCFdecon \cite{danelljan-CVPR16-adaptive}, CCOT \cite{martin-eccv16-beyond}, HDT \cite{qi-cvpr16-hdt}. We show the results of one-pass evaluation using the distance precision and overlap success rate in Figure \ref{fig:otb2015}. It indicates that our CREST tracker performs better than DCFs trackers HCFT and HDT with convolutional features. Overall, the CCOT method achieves the best result. Meanwhile, CREST achieves similar performance with DeepSRDCF in both distance precision and overlap threshold. Since there are more videos involved in this dataset in motion blur and fast motion, our CREST tracker is less effective in handling these sequences as CCOT.

{\flushleft \bf VOT-2016 Dataset.}
We compare our CREST tracker with state-of-the-art trackers on the VOT 2016 benchmark, including CCOT \cite{martin-eccv16-beyond}, Staple \cite{bertinetto-cvpr16-staple}, EBT \cite{zhu-cvpr16-beyond}, DeepSRDCF \cite{Danelljan-iccvw15-DeepSRDCF}, MDNet \cite{nam-cvpr16-mdnet} and SiamFC \cite{bertinetto-eccv16-fully}. As indicated in the VOT 2016 report \cite{kristan-eccvw16-vot}, the strict state-of-the-art bound is 0.251 under EAO metrics. For trackers whose EAO values exceed this bound, they will be considered as state-of-the-art trackers.

Table \ref{tab:vot} shows the results from our CREST tracker and the state-of-the-art trackers. Among these methods, CCOT achieves the best results under the EAO metric. Meanwhile, the performance of our CREST tracker is similar to those of Staple and EBT. In addition, these trackers perform better than DeepSRDCF, MDNet and SiamFC trackers. Note that in this dataset, MDNet does not use external tracking videos for training. According to the analysis of VOT report and the definition of the strict state-of-the-art bound, all these trackers can be regarded as state-of-the-art.

\subsubsection{Qualitative Evaluation}

\def\swthree{0.33\linewidth}
\def\swone{0.7\linewidth}
\begin{figure*}[t]
\begin{center}
\begin{tabular}{cc}
\begin{minipage}[t]{0.47\linewidth}
    \centering
    \renewcommand{\tabcolsep}{.1pt}
    \begin{tabular}{ccc}
    \vspace{-0.5mm}\includegraphics[width=\swthree]{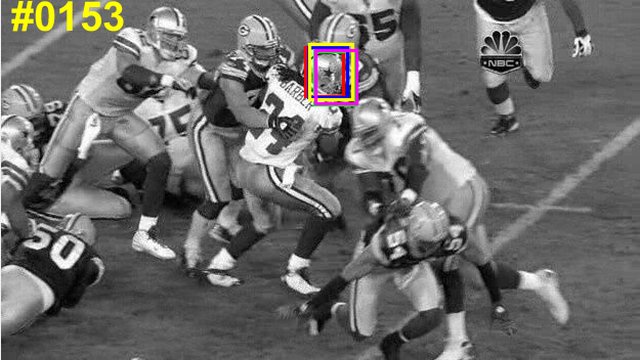}&
    \includegraphics[width=\swthree]{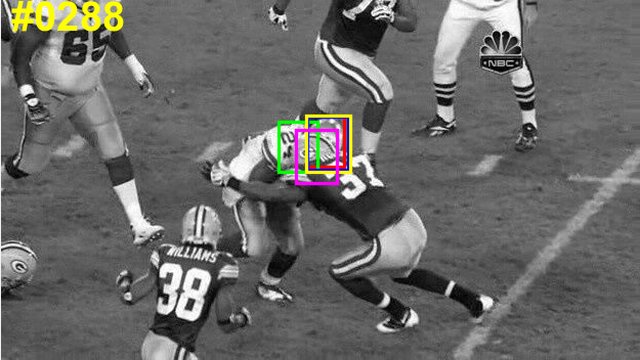}&
    \includegraphics[width=\swthree]{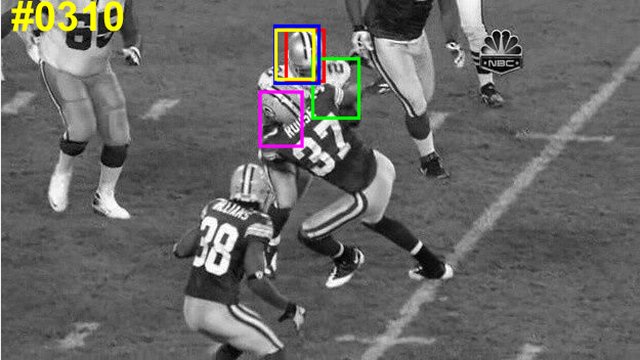}\\
    \vspace{-0.5mm}\includegraphics[width=\swthree]{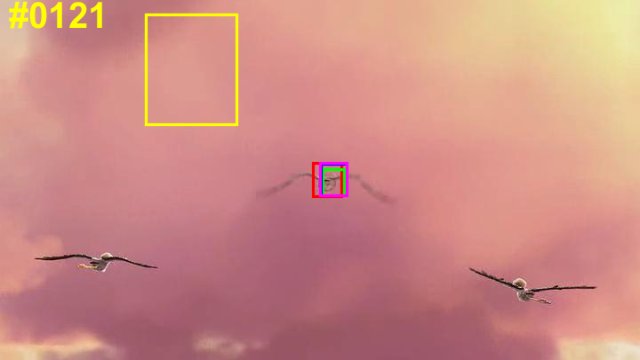}&
    \includegraphics[width=\swthree]{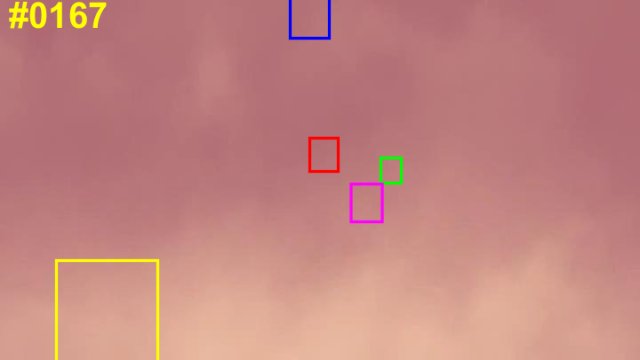}&
    \includegraphics[width=\swthree]{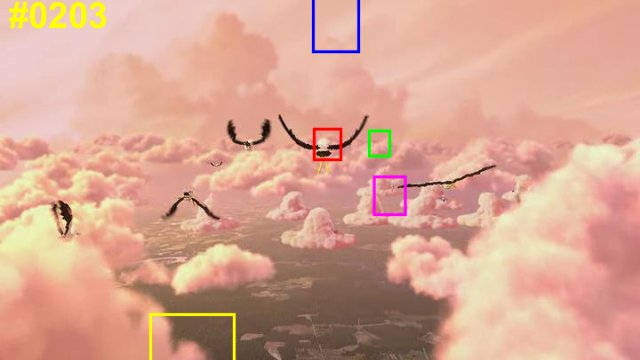}\\
    \vspace{-0.5mm}\includegraphics[width=\swthree]{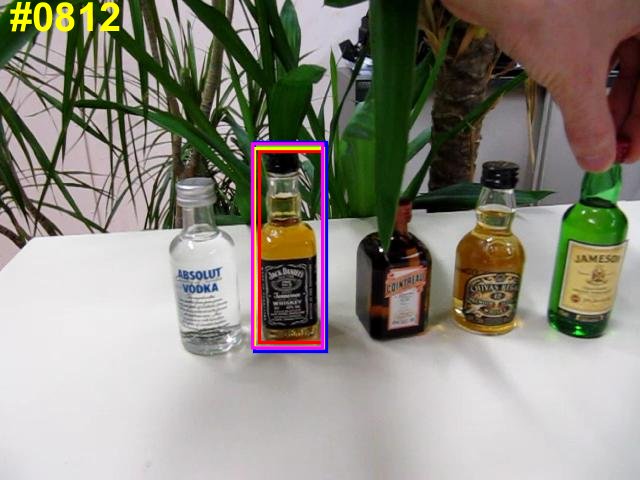}&
    \includegraphics[width=\swthree]{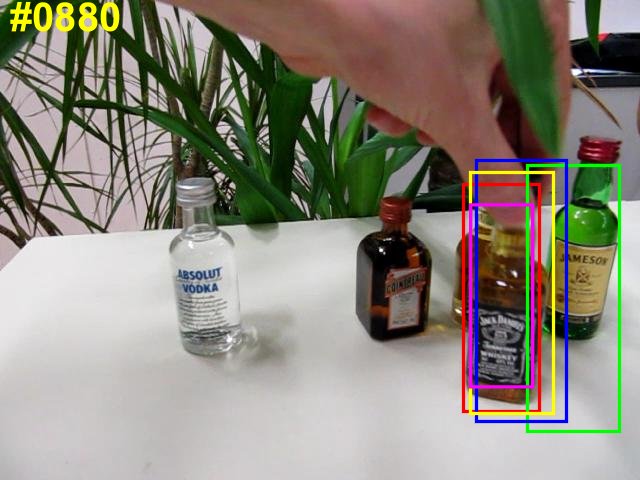}&
    \includegraphics[width=\swthree]{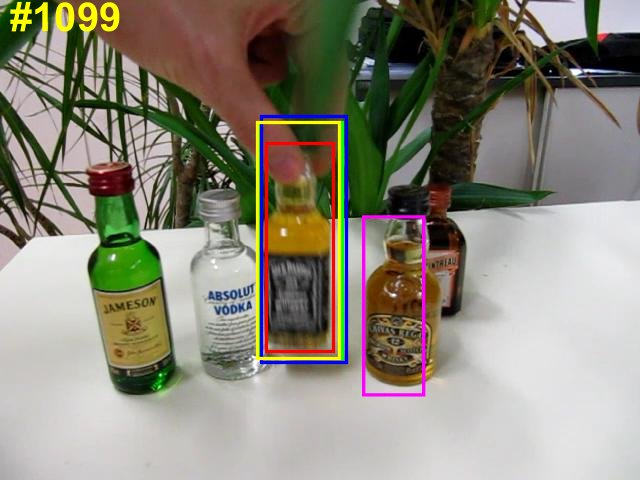}\\
    \vspace{-0.5mm}\includegraphics[width=\swthree]{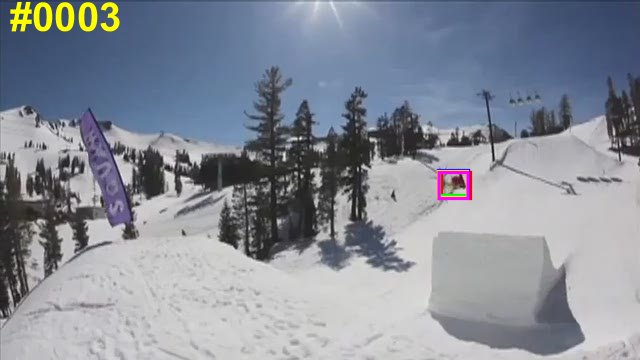}&
    \includegraphics[width=\swthree]{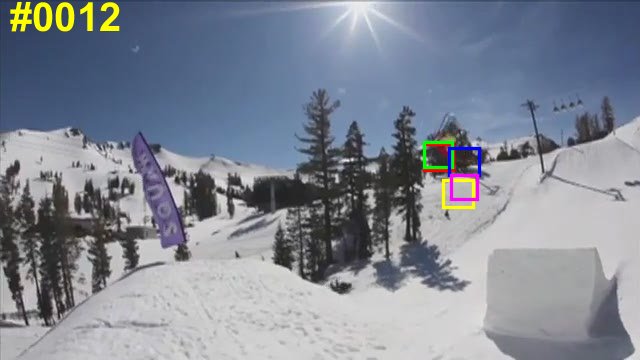}&
    \includegraphics[width=\swthree]{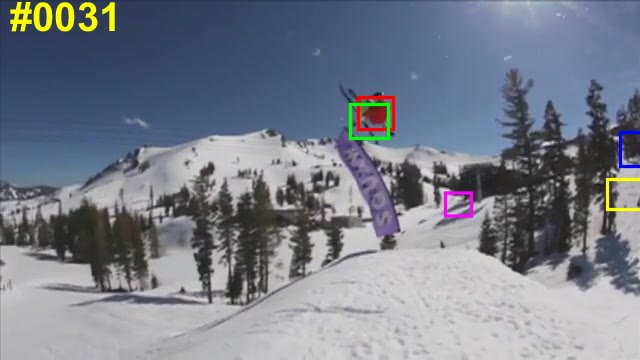}\\
    \vspace{-0.5mm}\includegraphics[width=\swthree]{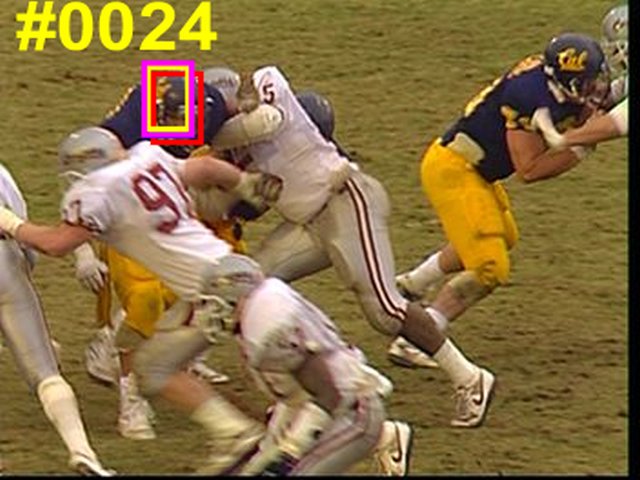}&
    \includegraphics[width=\swthree]{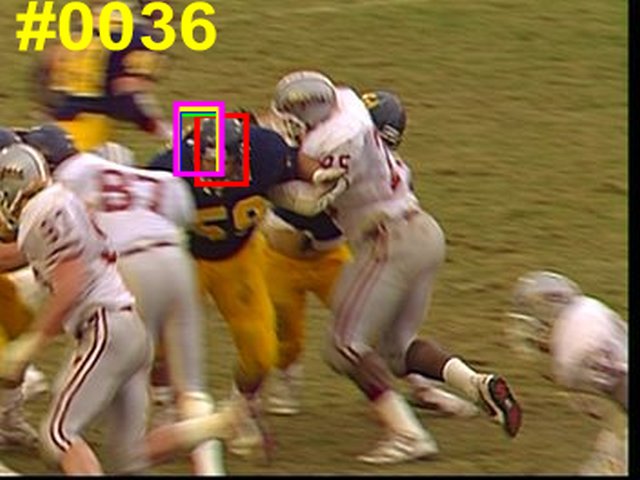}&
    \includegraphics[width=\swthree]{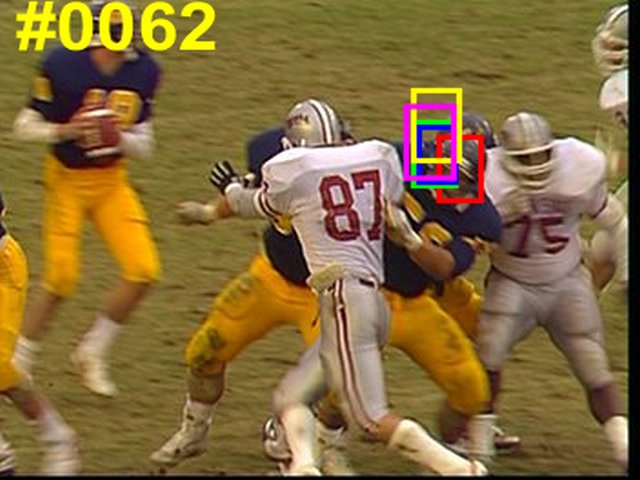}\\
    \includegraphics[width=\swthree]{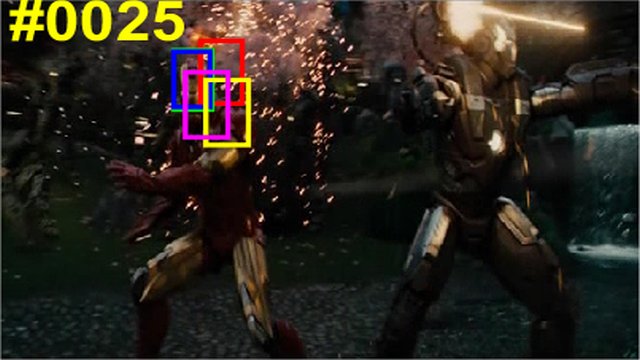}&
    \includegraphics[width=\swthree]{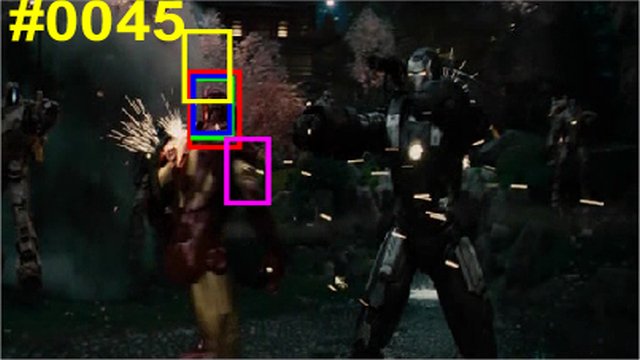}&
    \includegraphics[width=\swthree]{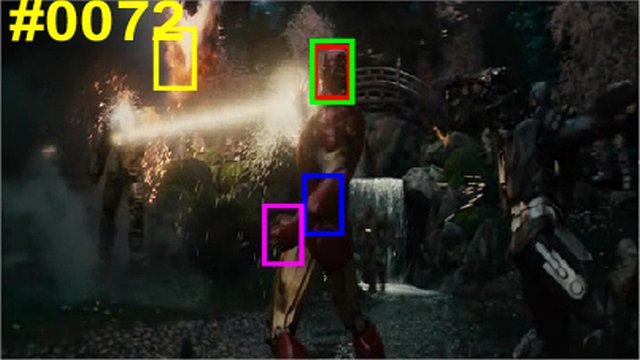}\\
    \end{tabular}
\end{minipage}
\begin{minipage}[t]{0.47\linewidth}
    \centering
    \renewcommand{\tabcolsep}{.1pt}
    \begin{tabular}{ccc}
    \vspace{-0.5mm}\includegraphics[width=\swthree]{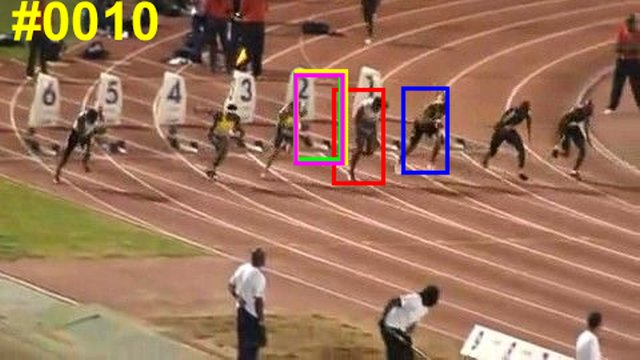}&
    \includegraphics[width=\swthree]{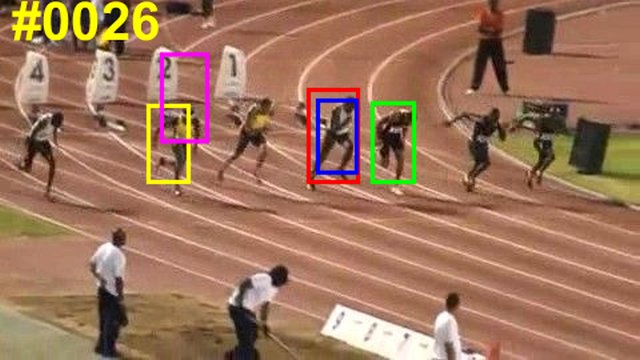}&
    \includegraphics[width=\swthree]{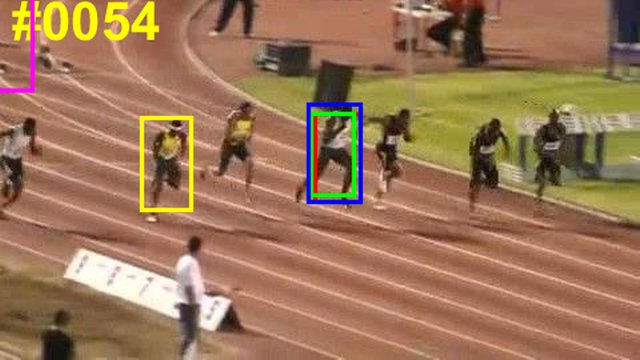}\\
    \vspace{-0.5mm}\includegraphics[width=\swthree]{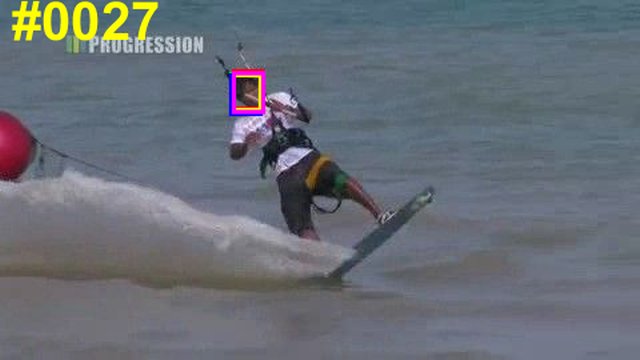}&
    \includegraphics[width=\swthree]{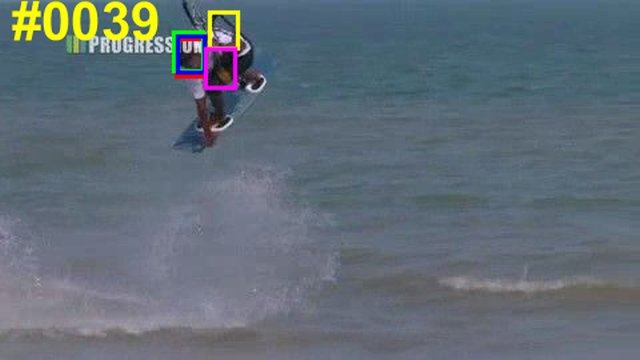}&
    \includegraphics[width=\swthree]{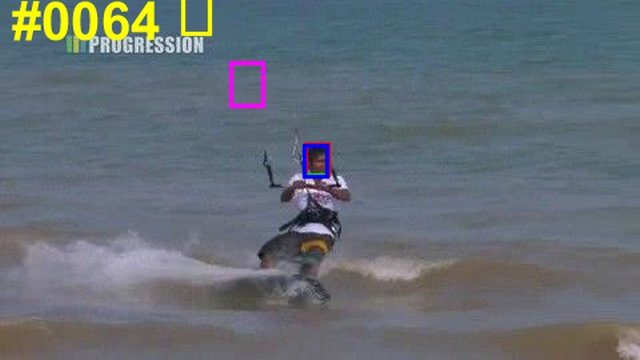}\\
    \vspace{-0.5mm}\includegraphics[width=\swthree]{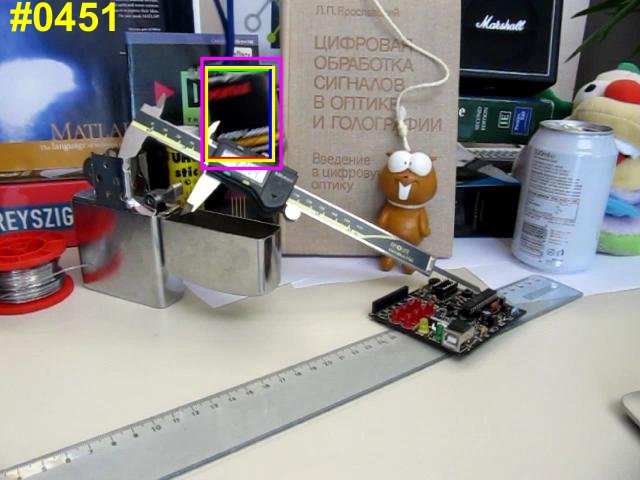}&
    \includegraphics[width=\swthree]{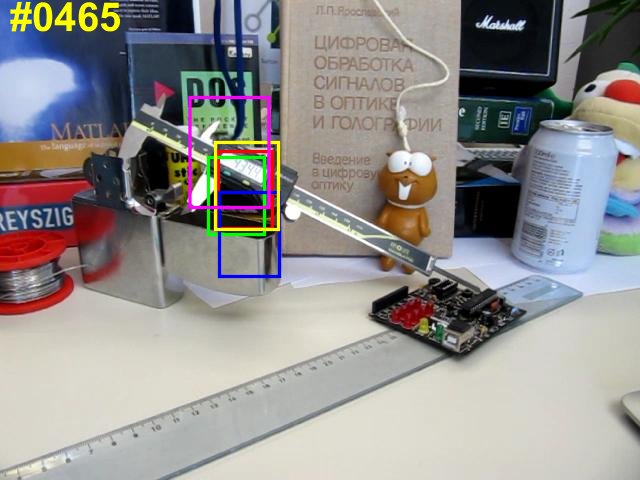}&
    \includegraphics[width=\swthree]{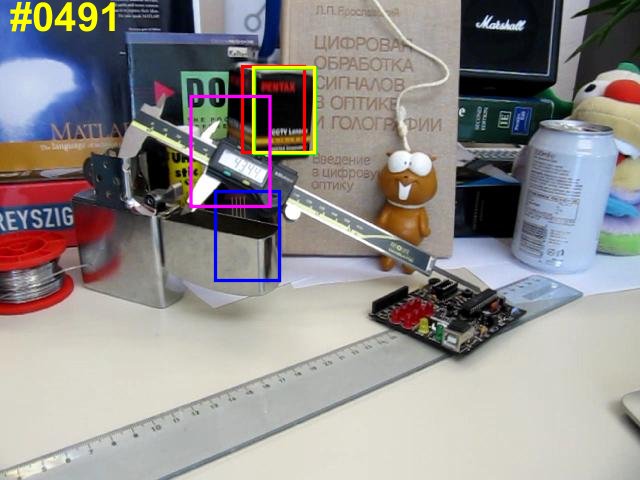}\\
    \vspace{-0.5mm}\includegraphics[width=\swthree]{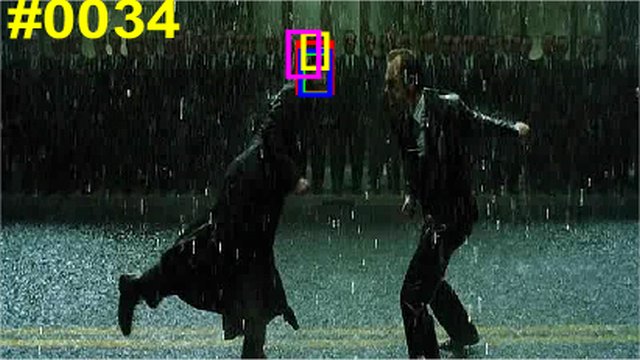}&
    \includegraphics[width=\swthree]{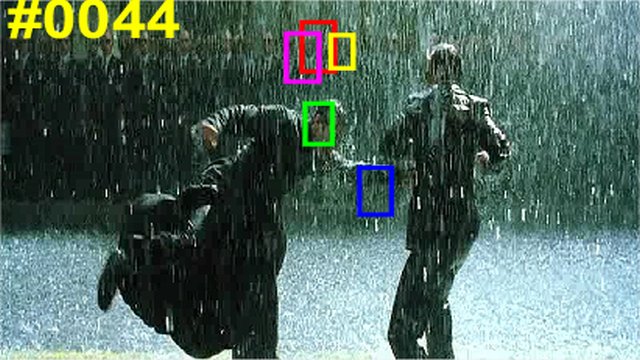}&
    \includegraphics[width=\swthree]{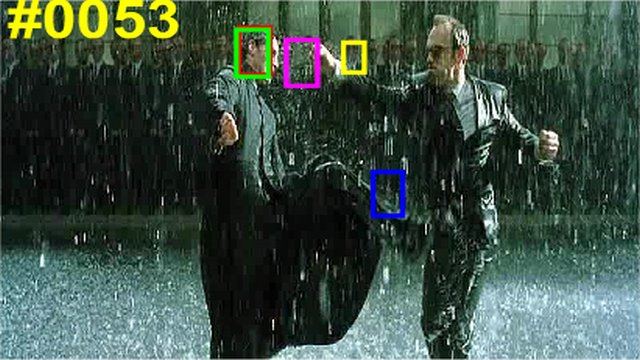}\\
    \vspace{-0.5mm}\includegraphics[width=\swthree]{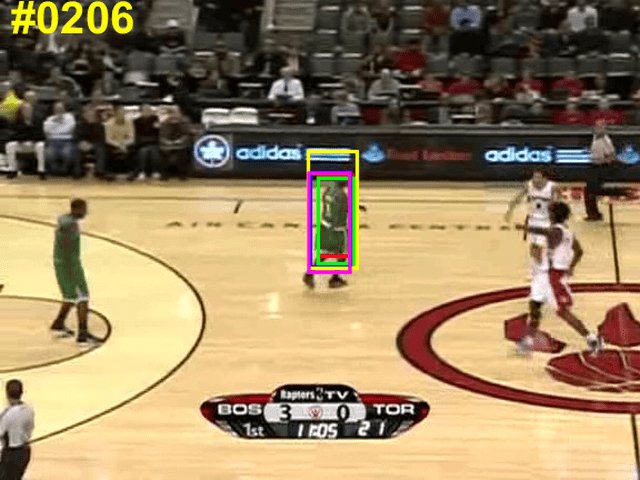}&
    \includegraphics[width=\swthree]{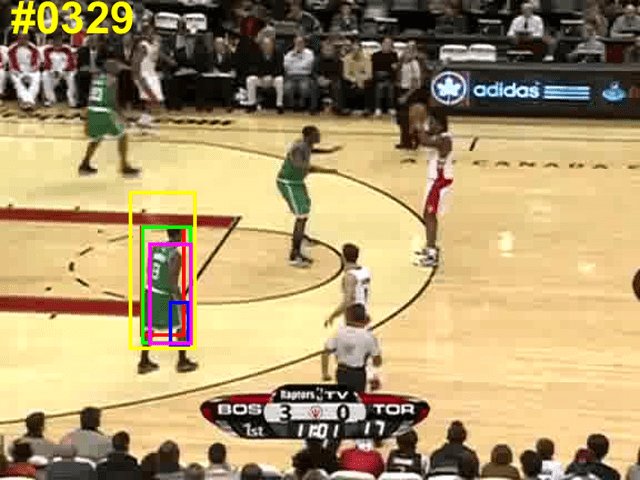}&
    \includegraphics[width=\swthree]{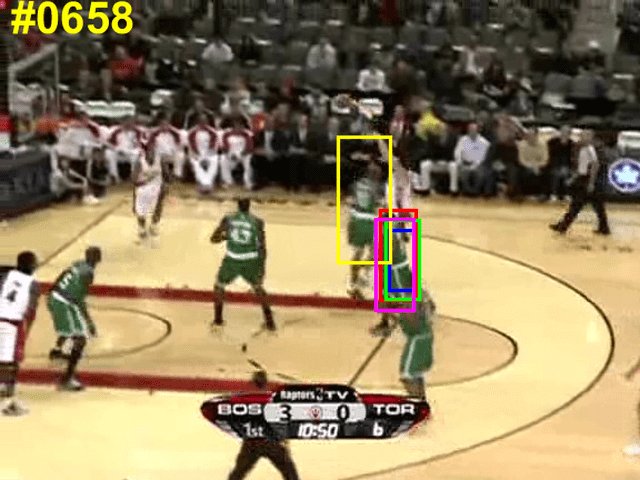}\\
    \includegraphics[width=\swthree]{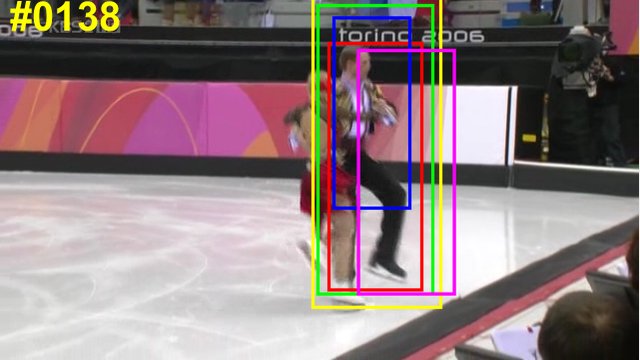}&
    \includegraphics[width=\swthree]{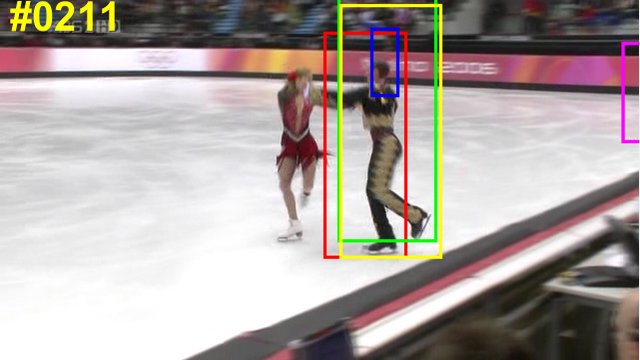}&
    \includegraphics[width=\swthree]{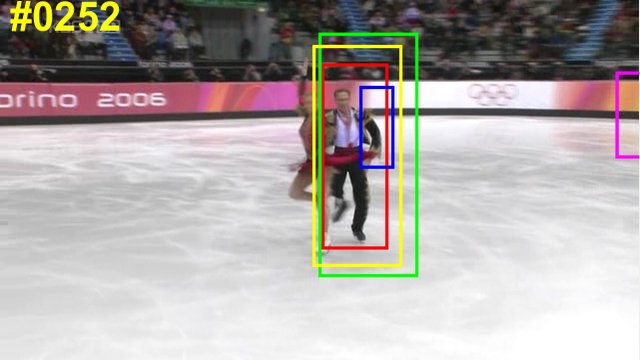}\\
    \end{tabular}
\end{minipage}\\
\end{tabular}
\begin{tabular}{c}
\includegraphics[width=\swone]{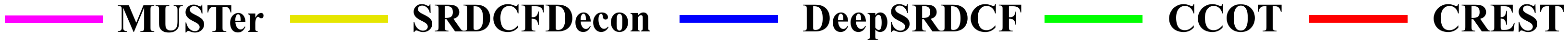}\\
\end{tabular}
\end{center}
\vspace{-5mm}
\caption{Qualitative evaluation of our CREST tracker, MUSTer \cite{hong-cvpr15-muster}, SRDCFDecon \cite{danelljan-CVPR16-adaptive}, DeepSRDCF \cite{Danelljan-iccvw15-DeepSRDCF}, CCOT \cite{martin-eccv16-beyond} on twelve challenging sequences (from left to right and top to down: \emph{Football}, \emph{Bolt2}, \emph{Bird1}, \emph{KiteSurf}, \emph{Liquor}, \emph{Box}, \emph{Skiing}, \emph{Matrix}, \emph{Football1}, \emph{Basketball}, \emph{Ironman} and \emph{Skating-2}, respectively). Our CREST tracker performs favorably against \ryn{the} state-of-the-art trackers.}
\label{fig:visual}
\end{figure*}

Figure \ref{fig:visual} shows some results of the top performing trackers: MUSTer \cite{hong-cvpr15-muster}, SRDCFDecon \cite{danelljan-CVPR16-adaptive}, DeepSRDCF \cite{Danelljan-iccvw15-DeepSRDCF}, CCOT \cite{martin-eccv16-beyond} and our CREST tracker on 12 challenging sequences. The MUSTer tracker does not perform well in all the presented sequences. This is because it adopts empirical operations for feature extraction (e.g., SIFT \cite{lowe-ijcv04-sift}). Although keypoint matching and long-term processing are involved, handcrafted features with limited performance are not able to differentiate the target and background. In comparison, SRDCFDecon incorporates the tracking-by-detection scheme to jointly optimize the model parameters and sample weights. It performs well on in-plane rotation (\emph{Box}), out-of-view (\emph{Liquor}) and deformation (\emph{Skating2}) because of detection. However, the sparse response leads to the limitation on background clutter (\emph{Matrix}), occlusion (\emph{Basketball}, \emph{Bird1}) and fast motion (\emph{Bolt2}). DeepSRDCF improves the performance through combining convolutional features with SRDCF. The dense prediction performs well on the fast motion (\emph{Bolt2}) and occlusion (\emph{Basketball}) scenes. However, the direct integration does not further exploit the model potential and thus limitation occurs on the illumination variation (\emph{Ironman}) and out-of-view (\emph{Bird1}). Instead of involving multiple convolutional features from different layers like CCOT, our CREST algorithm further exploits the model potential through DCF formulation and improves the performance through residual learning. We generate the target prediction in base layer while capturing the elusive residual via residual layers. These residual layers facilitate the learning process since they are jointly optimized with the base layer. As a result, the response map predicted by our network is more accurate for target localization, especially in the presented challenging scenarios. Overall, the visual evaluation indicates that our CREST tracker performs favorably against state-of-the-art trackers.

\section{Concluding Remarks}

In this paper, we propose CREST to formulate the correlation filter as a one-layer convolutional neural network named base layer. It integrates convolutional feature extraction, correlation response map generation and model update as a whole for end-to-end training and prediction. This convolutional layer is fully differentiable and allows the convolutional filters to be updated via online back propagation. Meanwhile, we exploit the residual learning to take target appearance changes into account. We develop spatiotemporal residual layers to capture the difference between the base layer output and the ground truth Gaussian label. They refine the response map through reducing the noisy values, which alleviate the model degradation limitations. Experiments on the standard benchmarks indicate that our CREST tracker performs favorably against state-of-the-art trackers.

\section{Acknowledgements}
\vspace{-1mm}
This work is supported in part by the NSF CAREER Grant \#1149783, gifts from Adobe and Nvidia.

\small
\bibliographystyle{ieee}
\bibliography{ref}

\end{document}